\documentclass[10pt,twocolumn,letterpaper]{article}

\usepackage[pagenumbers]{cvpr} % To force page numbers, e.g. for an arXiv version
\usepackage[dvipsnames]{xcolor}

\definecolor{cvprblue}{rgb}{0.21,0.49,0.74}
\usepackage[pagebackref,breaklinks,colorlinks,citecolor=cvprblue]{hyperref}

\usepackage[utf8]{inputenc} % allow utf-8 input
\usepackage[T1]{fontenc}    % use 8-bit T1 fonts
\usepackage{url}            % simple URL typesetting
\usepackage{booktabs}       % professional-quality tables
\usepackage{amsfonts}       % blackboard math symbols
\usepackage{nicefrac}       % compact symbols for 1/2, etc.
\usepackage{microtype}      % microtypography
\usepackage{xcolor}         % colors
\usepackage{comment}
\usepackage{amsmath}
\usepackage{graphicx}
\usepackage{caption}

\usepackage{subcaption}
\usepackage{makecell}    % for \thead
\usepackage[accsupp]{axessibility} % Improves PDF readability for those with visual impairments.

\newcommand{\draftonly}[1]{#1}
\renewcommand{\draftonly}[1]{}

\newcommand\Tstrut{\rule{0pt}{2.6ex}}         % = `top' strut
\newcommand\Bstrut{\rule[-0.9ex]{0pt}{0pt}}   % = `bottom' strut

\newcommand{\draftcomment}[1]{\draftonly{#1}}
\newcommand{\phillip}[1]{\draftcomment{\textcolor{blue}{\small [#1]$_{{PH}}$}}}

\newcommand{\anahita}[1]{\draftcomment{\textcolor{ForestGreen}{\small [#1]$_{{AB}}$}}}
\newcommand{\gustavo}[1]{\draftcomment{\textcolor{magenta}{\small [#1]$_{{GL}}$}}}
\newcommand{\tile}[1]{\draftcomment{\textcolor{magenta}{\small [#1]$_{{TL}}$}}}

\def\paperID{1516} % *** Enter the Paper ID here
\def\confName{CVPR}
\def\confYear{2024}

\title{SocialCounterfactuals: Probing and Mitigating Intersectional Social Biases in Vision-Language Models with Counterfactual Examples}

\author{Phillip Howard
\qquad
Avinash Madasu
\qquad
Tiep Le
\qquad
Gustavo Lujan Moreno\\
\qquad
Anahita Bhiwandiwalla
\qquad
Vasudev Lal \\\\
Intel Labs\\  \qquad {\tt\small \{phillip.r.howard, avinash.madasu, tiep.le, gustavo.lujan.moreno\}@intel.com}\\ \qquad {\tt\small \{anahita.bhiwandiwalla, vasudev.lal\}@intel.com}
}

\begin{document}

\maketitle
\begin{abstract}
While vision-language models (VLMs) have achieved remarkable performance improvements recently, there is growing evidence that these models also posses harmful biases with respect to social attributes such as gender and race. Prior studies have primarily focused on probing such bias attributes individually while ignoring biases associated with intersections between social attributes. This could be due to the difficulty of collecting an exhaustive set of image-text pairs for various combinations of social attributes. To address this challenge, we employ text-to-image diffusion models to produce counterfactual examples for probing intersectional social biases at scale. Our approach utilizes Stable Diffusion with cross attention control to produce sets of counterfactual image-text pairs that are highly similar in their depiction of a subject (e.g., a given occupation) while differing only in their depiction of intersectional social attributes (e.g., race \& gender). Through our over-generate-then-filter methodology, we produce SocialCounterfactuals, a high-quality dataset containing 171k image-text pairs for probing intersectional biases related to gender, race, and physical characteristics. We conduct extensive experiments to demonstrate the usefulness of our generated dataset for probing and mitigating intersectional social biases in state-of-the-art VLMs.
\end{abstract}    
\section{Introduction}
\label{sec:intro}

\begin{figure*}[h!]
    \centering
    \includegraphics[trim={2mm 2mm 2mm 
    2mm},clip,width=1\textwidth]{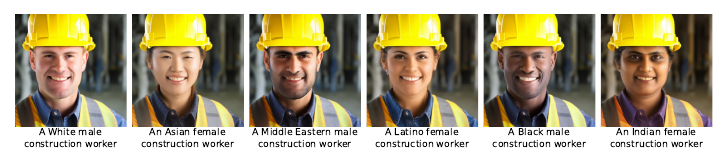}
    \caption{Examples of our counterfactual image-text pairs for probing intersectional race-gender bias in VLMs for the ``construction worker'' occupation. See Section~\ref{app:additional-examples} in Supplementary Material for additional examples.
    }
    \label{fig:main_examples}
\end{figure*}

Counterfactual examples, which study the impact on a response variable following a change to a causal feature, have proven valuable in natural language processing (NLP) for probing model biases and improving robustness to spurious correlation \citep{kaushik2019learning, gardner2020evaluating,wu2021polyjuice,wang2021robustness,yang2021exploring,fryer2022flexible,howard2022neurocounterfactuals}. While counterfactual examples for VLMs have been relatively unexplored, recent work  \citep{le2023coco} \phillip{this sentence is specifically referring to our COCO-Counterfactuals paper. I don't think there are any other similar papers to cite} has shown that text-to-image diffusion models with cross attention control can effectively produce multimodal counterfactual examples for VLM training, data augmentation and evaluation. This suggests that synthetic counterfactual examples generated by diffusion models could be an effective tool for probing and mitigating biases in VLMs.

Bias in pre-trained models can be viewed as spurious correlations, which are often attributed to the co-occurrence of non-causal (i.e., spurious) features with labels in datasets. During pre-training, models learn to exploit such correlations as shortcuts to achieving high in-domain performance on the training dataset \citep{geirhos2020shortcut}. Consequently, models which learn to rely on spurious correlations are more brittle and have worse out-of-domain (OOD) generalization \citep{singla2021salient, xiao2020noise}. 

Social biases are a particularly concerning type of spurious correlation learned by VLMs. Due to a lack of proportional representation for people of various races, genders, and other social attributes in image-text datasets \citep{birhane2021multimodal, zhao2021understanding,garcia2023uncurated}, VLMs learn biased associations between these attributes and various subjects (e.g., occupations). For example, given a gender- and race-neutral query such as ``A photo of an attorney'', a VLM may retrieve a disproportionate number of images of one particular race due to learned spurious correlations between this specific occupation and race. 

Prior studies on probing social biases in VLMs \citep{hall2023visogender,zhou2022vlstereoset,janghorbani2023multimodal,harrison2023run,hall2023vision} have primarily utilized real image-text pairs collected from existing datasets by identifying the co-occurrence of certain attributes with a target subject. However, this approach is limited by the availability of existing image-text pairs for various combinations of social attributes and subject types. Consequently, these prior studies \phillip{It's referring to the same prior studies that we reference earlier in the paragraph. I reworded it slightly to make that more clear} have focused exclusively on investigating biases associated with a single social attribute at a time while ignoring the potential role of intersectional bias (e.g., particular race-gender combinations) \citep{navigli2023biases}, which could be attributed to the difficultly of collecting an exhaustive set of image-text examples for various combinations of social attributes. Additionally, the large variability in how subjects can be naturally depicted in real images complicates the task of estimating bias in VLMs because disproportionate retrieval results could be attributed to other differences in images besides the social attribute. 

We overcome these limitations by leveraging text-to-image diffusion models to produce counterfactual image-text pairs for probing and mitigating social biases in VLMs (see Figure~\ref{fig:main_examples} and Section~\ref{app:additional-examples} in Supplementary Material for examples). Specifically, our approach utilizes Stable Diffusion \citep{rombach2021highresolution} with cross-attention control \citep{hertz2022prompt} to produce a set of highly similar counterfactual image-text examples which depict a common subject while differing only in intersectional social attributes. 
Text-to-image diffusion models are particularly well-suited for this task due to their ability to generate depictions of specific subjects with various combinations of different social attributes, which might be rare or missing from existing image-text datasets.
After generating candidate images, we apply three stages of filtering to ensure that only the highest-quality counterfactuals are retained.

We apply our methodology at scale to produce SocialCounterfactuals, an extensive dataset containing over 171k counterfactual image-text pairs for probing intersectional biases related to race, gender, and physical characteristics. To the best of our knowledge, SocialCounterfactuals is the largest resource released to-date for probing bias in VLMs and the only one which considers intersectional biases. Through extensive evaluation of six VLMs, we demonstrate its usefulness for uncovering intersectional social biases. Additionally, we conduct VLM training experiments using our dataset which demonstrate its ability to reduce skewness in retrieval results for social attributes. We make our dataset\footnote{Our dataset is available at \scriptsize\url{https://huggingface.co/datasets/Intel/SocialCounterfactuals}} and code\footnote{Our code is available at our \scriptsize\href{https://github.com/IntelLabs/multimodal_cognitive_ai/tree/main/SocialCounterfactuals}{GitHub repository}} publicly available.
\section{Related Work} 
\label{sec:related}

\subsection{Probing bias in pre-trained models}

Much of the prior work on probing social bias in pre-trained models has focused exclusively on language models, producing multiple datasets for measuring stereotypical bias along different social attributes, categories, demographic axes and stigmatized groups \citep{nadeem2020stereoset, nangia2020crows, smith2022m, mei2023bias}. Gender biases associated with pronoun and coreferences have also been extensively studied \citep{cao2019toward, webster2018mind, zhao2018gender, rudinger2018gender}. Some prior work has addressed topics related to intersectionality in bias evaluations \citep{kearns2018preventing,wang2022towards,tolbert2023correcting}. Additionally, various approaches for bias detection and mitigation for vision-only models have been proposed \citep{Wang_2020_CVPR, Iofinova_2023_CVPR, Aniraj_2023_ICCV, Brinkmann_2023_ICCV, fabbrizzi2022survey}.
%\citet{nadeem2020stereoset} introduced StereoSet, a natural English dataset for measuring stereotypical bias in language models for gender, race, professions, and religion. Concurrently, \citet{nangia2020crows} developed the the CrowS-Pairs dataset consisting of English sentences covering 9 bias categories. The HolisticBias dataset \citep{smith2022m} was subsequently proposed to investigate additional biases in language models, covering nearly 600 descriptor terms across 13 different demographic axes. \citet{mei2023bias} explored bias in masked language models across an even broader range of 93 stigmatized groups, which include conditions related to disease, disability, drug use, mental illness, religion, sexuality, socioeconomic status. These prior works focused exclusively on investigating bias in language models using text-only datasets, which differes from our focus on evaluating VLMs using paired image-text text samples. 

Approaches to generate synthetic datasets with fairness have been explored \cite{sattigeri2019fairness, friedrich2023fair, belgodere2023auditing, bhanot2021problem, lu2023machine}. For probing biases in VLMs, VLStereoSet \citep{zhou2022vlstereoset} extends the StereoSet dataset to the vision domain by sourcing images from Google search and using crowdsourced workers for annotation, resulting in a total of 1028 images. VisoGender \citep{hall2023visogender} consists of 690 manually-annotated images for benchmarking occupation-related gender bias in VLMs.
%collected from existing datasets and online search engines. 
The MultiModal Bias dataset \citep{janghorbani2023multimodal} evaluates bias in VLMs across 14 population subgroups and contains a total of 3800 human-annotated images obtained from Flickr. These datasets differ from ours primarily in their much smaller scale, their reliance on collection of data from existing sources \& human annotation, and their focus on investigating only a single attribute at a time (as opposed to our investigation of intersectional biases). Bias in VLMs used for text-to-image generation have also been investigated \citep{cho2023dall, wang2023t2iat, naik2023social}, but such studies differ from our focus on bias in image-text retrieval settings. 

\subsection{Mitigating bias in pre-trained VLMs}

A variety of methods have been proposed for mitigating the biases observed in pre-trained VLMs. These include adversarial approaches for prompt learning \citep{berg2022prompt}, fair sampling methods for reducing bias learned during training \citep{wang2021gender}, contrastive learning techniques for improving group robustness \citep{zhang2022contrastive}, learning additive residuals to offset image representations \citep{seth2023dear}, and eliminating biased directions in the text embedding space through projection matrices \citep{chuang2023debiasing}. 
% \citet{chuang2023debiasing} proposed a debiasing method for VLMs by projecting out biased directions in the text embedding space. Specifically, they estimate an orthogonal projection matrix using pairs of sentences with differing social attributes which can then be applied to models such as CLIP without additional training. 
\citet{smith2023balancing} introduced an approach for debiasing VLMs using synthetically-constructed contrast sets, 
% Similar to our work, they also utilize text-to-image diffusion models to produce synthetic examples for debiasing. However, they only 
which they use to produce 7946 image-text pairs for gender bias.
However, all of these prior works focus exclusively on debiasing models for a single social attribute at a time (e.g., gender) as opposed to our focus on debiasing for intersectional biases.

%whereas we generate much larger sets of counterfactual examples spanning various intersectional social attributes. Additionally, their GenSynth dataset contains only 7946 image-text pairs, whereas we apply our approach at scale to produce a dataset with over 233k image-text pairs.
\section{Generating SocialCounterfactuals}
\label{sec:method}

\begin{figure*}[h!]
    \centering
    \includegraphics[trim={2mm 2mm 2mm 
    2mm},clip,width=1\textwidth]{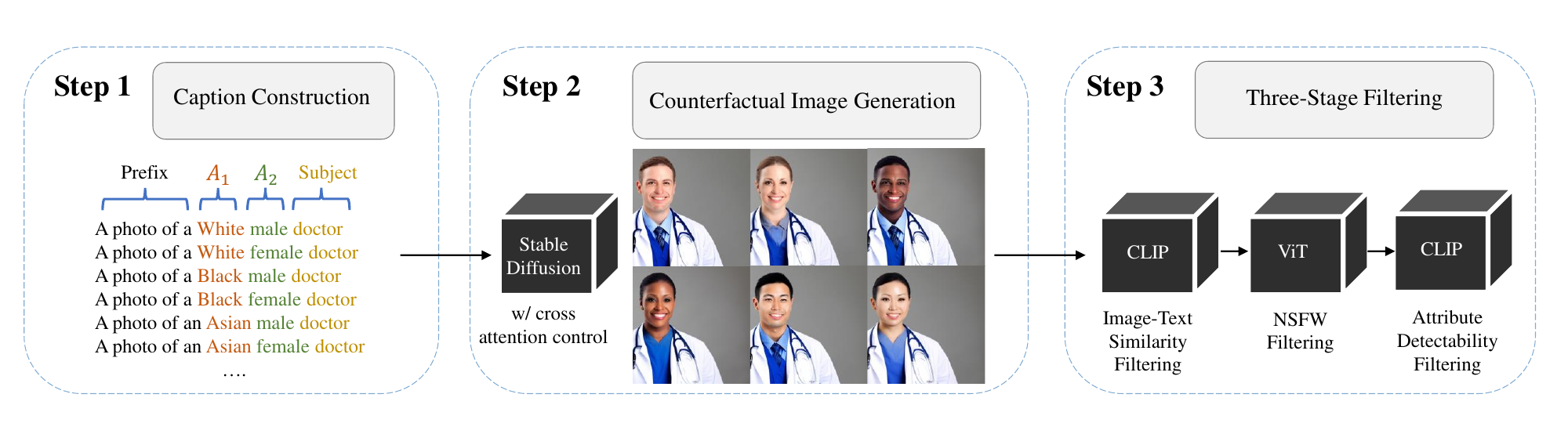}
    \caption{Overview of our methodology for generating SocialCounterfactuals.
    }
    \label{fig:method}
\end{figure*}

Our approach to creating counterfactual image-text examples for intersectional social biases consists of three steps. First, we construct sets of image captions describing a subject with counterfactual changes to intersecting social attributes. We then utilize a text-to-image diffusion model with cross attention control to over-generate sets of images corresponding to the counterfactual captions, where differences among images are isolated to the induced counterfactual change (i.e., the social attributes). Finally, we apply stringent filtering to identify only the highest-quality generations.

\subsection{Terminology}
In this work, we do not make any claims regarding gender identification or gender assignment, which we acknowledge to be unique to each individual regardless of their appearance or traits. We use \textit{perceived gender} as an inference made by a human annotator or model. We acknowledge that labels used in this study may differ from an individual's gender identity and could also vary based on different annotators' interpretations. We recognize that gender and gender identity is fluid and misjudgements in a binary paradigm could arise.

Similarly, we acknowledge that the six races we discuss in this paper - \textit{White, Black, Indian, Asian, Middle Eastern and Latino}, are not representative of all races.
We inherit this list of races from prior work \citep{karkkainen2019fairface}.
Any inference or reference to race and occupation is considered to be \textit{perceived race} and \textit{perceived occupation} (respectively), and does not aim to associate any bias with any groups of individuals.

\subsection{Constructing captions for probing social biases}
\label{sec:captions}

Consider the task of creating a caption $C^{s}_{p, a_{1},a_{2}}$ beginning with prefix $p$ and describing a subject $s$ which possesses a pair of attributes $a_{1}$ and $a_{2}$\footnote{See Section~\ref{app:three-attributes} for discussion of extension to more than two attributes}. Given a set of prefixes $P$, a set of subjects $S$, and attribute sets $A_{1}, A_{2}, ..., A_{k}$, we populate the following template to obtain our captions:
\begin{equation*}
\begin{aligned}
    & C^{s}_{p, a_{1},a_{2}} = \textrm{<$p$> <$a_{1}$> <$a_{2}$> <$s$>} \\
    & \forall \ p \in P, s \in S, a_{1} \in A_{i}, a_{2} \in A_{j}, (i, j) \in \{1, ..., k \ | \ i \ne j\} \\
    % \textrm{A photo of a <$a_{1}$> <$a_{2}$> <$s$>} \\
    % \textrm{A picture of a <$a_{1}$> <$a_{2}$> <$s$>} \\
    % \textrm{An image of a <$a_{1}$> <$a_{2}$> <$s$>} \\
\end{aligned}
\end{equation*}
% For example, given the prefix $p=\textit{A photo of a}$, subject $s=\textit{doctor}$, and attributes $a_{1}=\textit{Hispanic}, a_{2}=\textit{female}$, we populate the caption $C^{s}_{p, a_{1},a_{2}} = \textit{A photo of a Hispanic female doctor}$. 
For example, given the prefix \textit{A photo of a}, the subject \textit{doctor}, and attribute pair $(\textit{Asian}, \textit{Female})$, we construct the caption \textit{A photo of an Asian female doctor}.
We construct captions in this manner using a set of occupations as our subjects and three sets of attributes for measuring social bias (gender, race, and physical characteristics). 
Captions are grouped into counterfactual sets, where each set contains all captions corresponding to a given prefix and subject. Using 260 occupations, 4 prefixes, 6 races, and 5 physical characteristics, and 2 gender terms\footnote{We also explored the generation of other subjects and social attributes with our method. See Section~\ref{app:other-subjects-and-attributes} of Supplementary Material for details.}, we produced a total of 54,080 captions which were grouped into 3,120 counterfactual sets (see Section~\ref{app:caption-details} of Supp. Material for additional details and examples). While we categorize these social attributes with the aim of probing biases, we recognize the limitations inherent to this process and acknowledge that attributes such as gender and race are not considered by all individuals to exist as discrete categories (see Section \ref{sec:limitations} for additional discussion). 

\subsection{Counterfactual image generation}

After generating sets of counterfactual captions, we use text-to-image diffusion models to produce images for each caption. In order to precisely measure the impact of social attribute differences, it is desirable for images within a counterfactual set to only differ in how the social attributes differ across captions. However, this is challenging for diffusion models as even minor changes to a prompt can result in the generation of images with significant differences. For example, changing the attributes \textit{Hispanic female} to \textit{Asian male} in the prompt \textit{A photo of a Hispanic female doctor} may produce other undesired modifications to the image that extend beyond the induced counterfactual change (e.g., changes to the background). This complicates the task of quantifying the impact of model bias attributed to the changed social attributes on retrieval results, as other differences between the generated images could contribute to a VLM's preference for retrieving particular images.

\citet{hertz2022prompt} proposed Prompt-to-Prompt to address this issue by injecting cross-attention maps during denoising steps to control attention between certain pixels and tokens, which enables separate generations to maintain many of the same details while isolating differences to how the text prompts differ. %An example of counterfactual image-text pairs $(C_{o},I_{o}^{s})$ and $(C_{c},I_{c}^{s})$ generated with and without prompt-to-prompt is shown in Figure~\ref{fig:p2p_example}, illustrating how Prompt-to-Prompt enables the principle of minimal text edits for NLP counterfactuals to be extended to image generation. 
% \begin{figure}[t!]
%     \centering
%     \includegraphics[trim={2mm 2mm 2mm 2
%     2mm},clip,width=1\columnwidth]{images/p2p_example.pdf}
%     \caption{Examples of \cococounterfactuals generated with Prompt-to-Prompt (left) and without (right). Prompt-to-prompt enables us to extend the principle of minimal-edit text counterfactuals to the visual domain, isolating image differences to only the changed causal feature. 
%     }
%     \label{fig:p2p_example}
% \end{figure}
However, \citet{brooks2023instructpix2pix} noted that some changes require varying the parameter $p$ in Prompt-to-Prompt, which controls the number of denoising steps with shared attention weights. 
For example, changes that require more substantial structural modifications to the image may necessitate less overall similarity between the resulting images and thus fewer shared attention weights. 
We therefore adopt their proposed approach of over-generating 100 image pairs with Prompt-to-Prompt by sampling $p \sim U(0.1,0.9)$, which we then subsequently filter to retain only the highest-quality generated candidates (Section~\ref{sec:filtering}). 

We extend Prompt-to-Prompt for image pairs from \citet{brooks2023instructpix2pix} to support batched generation of multiple images with cross attention control. This enables simultaneous generation of entire sets of counterfactual images which differ only according to the social attribute differences across prompts (e.g., Figure~\ref{fig:main_examples}).
In total, we over-generate 5,408,000 images for 54,080 captions.\gustavo{I changed these numbers to match Table \ref{table:filtering-stats}. Recheck once we decided either to include/exclude duplicates}

% After filtering with CLIP, we keep up to 10 of the highest-scoring counterfactual sets according to CLIP directional similarity.%, resulting in a dataset comprised of $x$ counterfactual sets with a total of $y$ images. \phillip{todo: calculate final dataset statistics and add here}

\subsection{Three-Stage Filtering}
\label{sec:filtering}

\paragraph{CLIP image-text similarity filtering.} After over-generating 100 candidate image sets for each of our templates, we first filter the candidates using CLIP \citep{radford2021learning} to ensure a minimum cosine similarity of 0.2 between the encoding of each caption and its corresponding generated image. We also apply a similar filtering criteria between pairs of images in each set, requiring the cosine similarity of CLIP image encodings within the set to be greater than 0.2. These filtering criteria help ensure that images accurately depict the subjects described in each caption while also retaining a high-degree of similarity to each other, thereby ensuring that they represent valid counterfactual examples.
% , with the best image pair chosen according to the directional similarity in CLIP space \citep{gal2022stylegan}
% \phillip{Add equation to appendix?}.
% \begin{equation}
%     CLIP_{dir} = \frac{(E_{T}(C_{c}) - E_{T}(C_{o})) \cdot (E_{I}(I_{c}^{s}) - E_{I}(I_{o}^{s}))}{||E_{T}(C_{c}) - E_{T}(C_{o})|| \ ||E_{I}(I_{c}^{s}) - E_{I}(I_{o}^{s})||}
% \end{equation}
% where $E_{T}$ and $E_{I}$ are CLIP's text and image encoders (respectively). The $CLIP_{dir}$ metric measures the consistency in changes between the two images $(I_{o}^{s},I_{c}^{s})$ and their corresponding captions $(C_{o},C_{c})$. Thus, selecting images with a higher $CLIP_{dir}$ improves the overall quality of our generated counterfactuals via greater consistency between the alterations made in both modalities. 

\paragraph{NSFW Filtering with ViT.}

Manual examination of generated images revealed instances of not-safe-for-work (NSFW) content. We therefore applied a NSFW filter \footnote{\scriptsize\url{https://huggingface.co/Falconsai/nsfw_image_detection}} which uses a fine-tuned vision transformer (ViT) for NSFW image classification. This filter removes 0.9-2.7\% of generated images (see Table \ref{table:filtering-stats} in Supplementary Material for details).
% 0.9\% of the images were removed for the (race, gender) attribute type, 1.4\% for (Physical Characteristics, Gender) and 2.7\% for (Physical Characteristics, Race).
\gustavo{These numbers come from Table \ref{table:filtering-stats}. Recheck once we decided either to include/exclude duplicates}

\paragraph{CLIP Attribute Detectability Filtering.}
To further ensure data quality, we filter counterfactual sets based on CLIP's ability to discern targeted social attributes in each image using a two-phase approach (see Section~\ref{app:dataset-filtering-detail} in Supp. Material for additional details).
In the first phase, we randomly sampled 100 counterfactual sets for each pair of attribute types.
%, says $S_{(A_i, A_j)}$. 
For each attribute type,
%$A \in [A_i, A_j]$, 
we then manually labeled the sampled counterfactual sets according to whether or not they should be filtered out based on a lack of detectability of the targeted attribute. Specifically, we label how many images in a counterfactual set possess their targeted attribute.
% \footnote{We acknowledge that in spite our best efforts, it is possible this manual analysis could propagate the annotator's bias for these social attributes.}
%$a \in A$ 
% discernible by a human annotator. Such threshold-based heuristics were manually determined 
%for each attribute type in each pairs of  attribute types 
% to guarantee counterfactual sets remaining after filtering have reasonably-high quality.

In the second phase, we develop a \emph{learnable threshold}-based heuristics to automatically label whether a counterfactual set should be filtered with respect to an individual attribute type. These threshold-based heuristics are applied according to how many of a set's constituent images have their respective targeted attributes discernible by CLIP-based image-text similarity scores (rather than by a human annotator). Thresholds were heuristically derived to obtain high correspondence between automatic filtering with CLIP and those filtered by the manual human annotation.
% accuracy with respect to their corresponding labels manually constructed from the first phase for the same randomly sampled 100 counterfactual sets.
This process produces a separate learned threshold for each attribute type and attribute type pair combination. A counterfactual set is filtered out due to a lack of detectability for an attribute type if the number of its constituent images whose corresponding targeted attribute is discernible by CLIP is less than the corresponding learned threshold. 

To estimate the quality of our generated dataset and the impact of filtering, we randomly sampled 100 counterfactual sets depicting Race-Gender intersectional social attributes.
Prior to CLIP attribute detectability filtering, we found that 90.8\% of the images accurately depicted their corresponding captions.
Applying attribute detectability filtering further increases this figure to 97.5\%, which demonstrates the value of our filtering methodology and the high quality it ensures in our dataset (see Section~\ref{sec:error-analysis} of Supp. Material for details).
% Table \ref{table:error-analysis} shows that before the detectability filter only 90.8\% of the images were good for the (race, gender) pair. The detectability filter on race yielded 0.99 and 0.96 precision for male and female respectively. After applying the filter the correct detection for both male and female produced 97.5\% of good images.

Table~\ref{tab:dataset-details} provides details on the total number of counterfactual sets and images which remain in our dataset after filtering. We group counterfactual sets into three dataset segments based on the pair of attribute types used to construct the captions, which are detailed in each row of Table~\ref{tab:dataset-details}. In total, our dataset consists of 13,824 counterfactual sets with 170,832 \gustavo{I just changed this number to reflect updated info from \ref{tab:dataset-details}} image-text pairs, which represents the largest paired image-text dataset for investigating social biases to-date.

\begin{table}
\footnotesize
\begin{center}
\resizebox{1\columnwidth}{!}
{
\begin{tabular}{l c c c } %
\toprule
Attribute Pair & Counterfactual Sets & Images Per Set & Total Images\\
\midrule

$(\textrm{Race}, \textrm{Gender})$ & 7,936 & 12 & 95,232 \\
$(\textrm{Physical Char.}, \textrm{Gender})$ & 5,052 & 10 & 50,520 \\
$(\textrm{Physical Char.}, \textrm{Race})$ & 836 & 30 & 25,080\\
\bottomrule
\end{tabular}
}
\caption{Details of the number of counterfactual sets, images per set, and total images which remain in our dataset after filtering}
\label{tab:dataset-details}
\end{center}
\end{table}

\phillip{Need to mention human evaluation results before and after the attribute detectability filtering}

% \begin{table}
% \footnotesize
% \begin{center}
% \resizebox{1\columnwidth}{!}
% {
% \begin{tabular}{l c c c } %
% \toprule
% Attribute Pair & Counterfactual Sets & Images Per Set & Total Images\\
% \midrule

% $(\textrm{Race}, \textrm{Gender})$ & 10,064 & 12 & 120,768 \\
% $(\textrm{Physical Char.}, \textrm{Gender})$ & 7,571 & 10 & 75,710 \\
% $(\textrm{Physical Char.}, \textrm{Race})$ & 1,232 & 30 & 36,960\\
% \bottomrule
% \end{tabular}
% }
% \caption{Details of the number of counterfactual sets, images per set, and total images which remain in our dataset after filtering}
% \label{tab:dataset-details}
% \end{center}
% \end{table}
\section{Probing Intersectional Biases}
\label{sec:probing}

To probe intersectional social biases in VLMs, we calculate MaxSkew over our dataset.\footnote{We also provide results for other evaluation metrics in Section~\ref{app:ndkl-results}}
% % \footnote{We also provide results for normalized discounted cumulative KL-divergence (NDKL) in the appendix}
We describe this this metric in Section~\ref{sec:metrics} and detail our evaluation results in Section~\ref{sec:results}.\footnote{See Section~\ref{app:probing-additional-reults} for additional details and probing results}

\subsection{Evaluation Metrics}
\phillip{add NDKL?}

\begin{figure*}[t!]
    \centering
    \begin{subfigure}[b]{0.33\textwidth}
    \includegraphics[trim={2mm 2mm 2mm 
    2mm},clip,width=1\columnwidth]{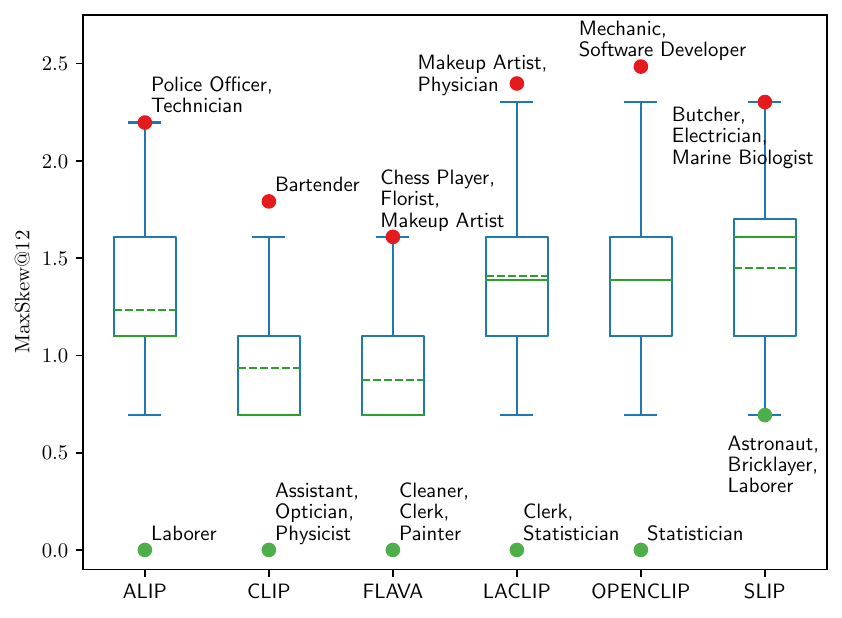}
    \caption{Race-Gender}
    \label{fig:race-gender}
    \end{subfigure}
    \begin{subfigure}[b]{0.33\textwidth}
    \includegraphics[trim={2mm 2mm 2mm 
    2mm},clip,width=1\columnwidth]{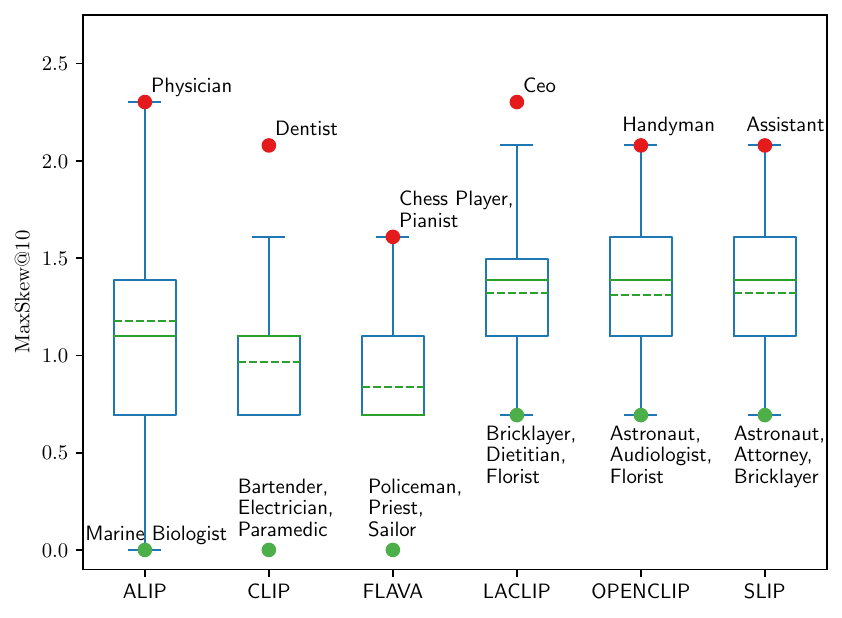}
    \caption{Physical Characteristics-Gender}
    \label{fig:physical-gender}
    \end{subfigure}
    \begin{subfigure}[b]{0.33\textwidth}
    \includegraphics[trim={2mm 2mm 2mm 
    2mm},clip,width=1\columnwidth]{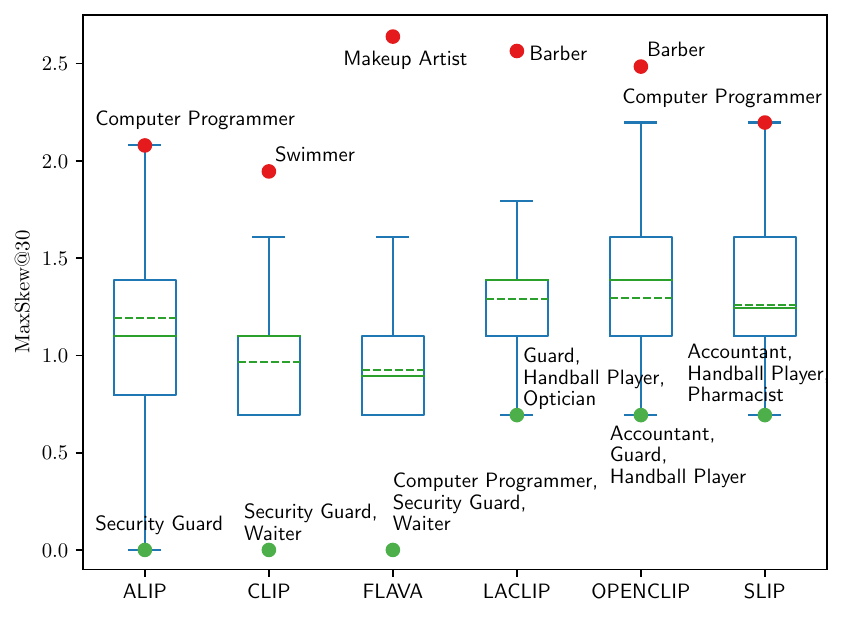}
    \caption{Physical Characteristics-Race}
    \label{fig:physical-race}
    \end{subfigure}
    \caption{
    Distribution of $\textrm{MaxSkew}@K$ measured across occupations for (a) Race-Gender, (b) Physical Characteristics-Gender, and (c) Physical Characteristics-Race intersectional biases. Max (min) values are plotted as red (green) circles with corresponding occupation names
    }
    \label{fig:maxskew}
\end{figure*}

% \subsection{Evaluation metrics}
\label{sec:metrics}

Let $q$ denote a text query and $R_{K}(q)$ denote the set of top-$K$ ranked images retrieved by a VLM for $q$. For a given attribute pair $(a_{i}, a_{j})$, we denote the desired proportion $d$ of retrieved images with the corresponding attributes as $p_{d(q), (a_{i}, a_{j})}$ and the actual proportion as $p_{R_{K}(q), (a_{i}, a_{j})}$. \citet{geyik2019fairness} define $\textrm{Skew}@K$ for attributes $(a_{i}, a_{j})$ in retrieval results $R_{K}(q)$ as:
\begin{equation*}
    \textrm{Skew}_{(a_{i},a_{j})}@K(R_{K}(q)) = \log(\frac{p_{R_{K}(q), (a_{i}, a_{j})}}{p_{d(q), (a_{i}, a_{j})}})
\end{equation*}
In essence, $\textrm{Skew}@K$ measures the ratio of the proportion of top-$K$ retrieved images having a set of attributes to the desired proportion. To aggregate $\textrm{Skew}@K$ over the various attributes under consideration, \citet{geyik2019fairness} further proposed the following $\textrm{MaxSkew}@K$ metric:
\begin{equation*}
    \textrm{MaxSkew}@K(R_{K}(q)) = \max_{(a_{i}, a_{j}) \in A} \textrm{Skew}_{(a_{i},a_{j})}@K(R_{K}(q))
\end{equation*}
where $A$ denotes the set of all attribute pairs. We calculate $\textrm{MaxSkew}@K$ by retrieving images from our counterfactual sets using prompts which are neutral with respect to the investigated attributes. For example, given a prompt constructed from the template ``A <race> <gender> construction worker'' (Figure~\ref{fig:main_examples}), we form its corresponding attribute-neutral prompt ``A construction worker.'' 

We construct neutral prompts in this manner for each unique combination of prefixes and subjects, averaging their text representations across different prefixes to obtain a single text embedding for each subject. $\textrm{Skew}@K$ and $\textrm{MaxSkew}@K$ are then calculated by retrieving the top-$K$ images for the computed text embedding from the set of all images generated for the subject which met our filtering and selection criteria. We set $K = |A_{1}| \times |A_{2}|$, where $A_{1}$ and $A_{2}$ are the investigated attribute sets.
% \footnote{We set $K$ equal to the number of unique attribute combinations in the evaluated counterfactual sets}.

\subsection{Results}
\label{sec:results}

\begin{figure}[t!]
    \centering
    \includegraphics[trim={2mm 2mm 2mm 
    2mm},clip,width=1\columnwidth]{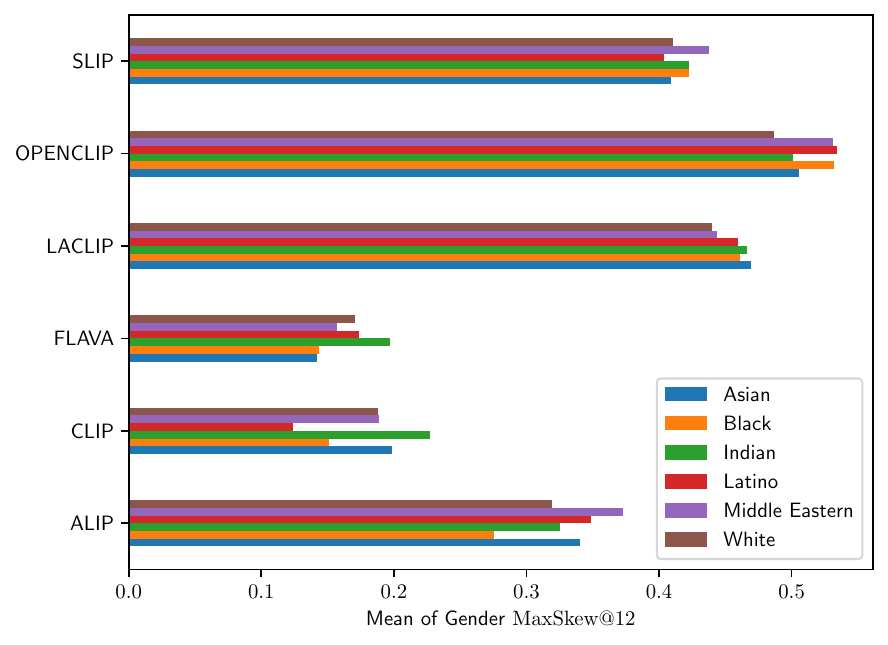}
    \caption{
    Mean of (marginal) gender $\textrm{MaxSkew}@K$ measured across occupations for different races.
    }
    \label{fig:maxskew-by-race}
\end{figure}

\begin{figure*}[t!]
    \centering
    \includegraphics[trim={2mm 2mm 2mm 
    2mm},clip,width=1\textwidth]{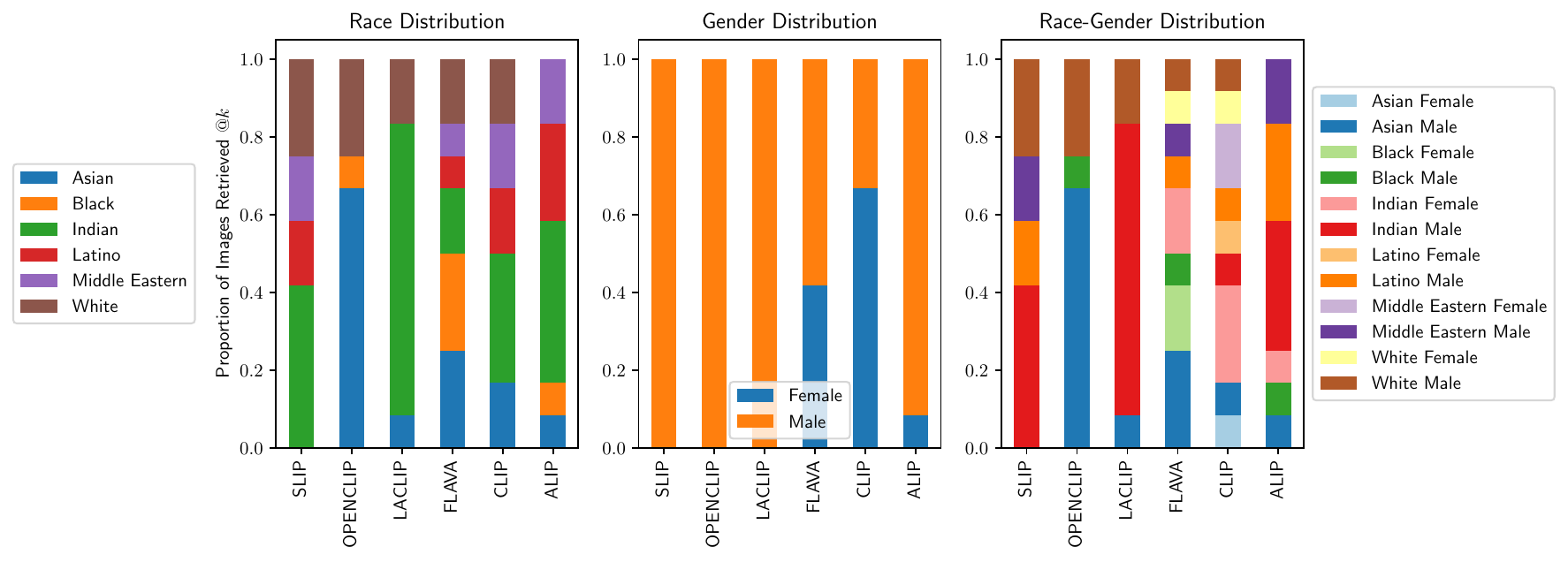}
    \caption{
    Proportion of images retrieved $@k=12$ using neutral prompts for the `Doctor' occupation.
    }
    \label{fig:doctor}
\end{figure*}

Figure~\ref{fig:maxskew} provides boxplots of the intersectional bias $\textrm{MaxSkew}@K$ distribution for six state-of-the-art VLMs: ALIP \citep{yang2023alip}, CLIP \citep{radford2021learning}, FLAVA \citep{singh2022flava}, LaCLIP \citep{fan2023improving}, OpenCLIP \citep{cherti2023reproducible}, and SLIP \citep{mu2022slip}.
We measure the $\textrm{MaxSkew}@K$ distribution across occupations separately using the three segments of our dataset for Race-Gender (Figure~\ref{fig:race-gender}), Physical Characteristics-Gender (Figure~\ref{fig:physical-gender}), and Physical Characteristics-Race (Figure~\ref{fig:physical-race}) intersectional biases.
All six evaluated VLMs exhibit significant skewness in retrieval for attribute-neutral occupation prompts, with CLIP and FLAVA having the lowest overall $\textrm{MaxSkew}@K$. 
Among the three segments of our dataset, Physical Characteristics-Gender intersectional biases tend to have lower $\textrm{MaxSkew}@K$ scores while Race-Gender intersectional biases have the greatest amount of skewness. 
% Notably, ALIP exhibited zero skewness for only two occupations (`Laborer' and `Swimmer'), CLIP exhibited zero skewness for only one occupation (`Engineer'), and SLIP exhibited positive skewness across all occupations.

In addition to illustrating the overall distribution of $\textrm{MaxSkew}@K$ across occupations, the boxplots in Figure~\ref{fig:maxskew} also provides the occupation names for minimum values (green circles) and maximum values (red circles).
These labels show some commonalities across models, such as FLAVA and LaCLIP both having their greatest Race-Gender $\textrm{MaxSkew}@K$ values for the `Makeup Artist' profession. 
We also observe that  LaCLIP and OpenCLIP have their greatest skewness in retrieval from the Physical Characteristics-Race segment for the `Barber' occupation. 
Notably, both CLIP and FLAVA have multiple occupations with zero skewness across all segments of the dataset, while SLIP has no occupations with zero skewness across the three segments.

Our dataset can also be used to evaluate individual (i.e., marginal) bias for a specific attribute type by filtering on the value of the other attribute depicted in a counterfactual set. $\textrm{MaxSkew}@K$ can then be calculated as previously described to estimate bias in retrieval results for a single social attribute at a time. To illustrate, we estimated the marginal gender bias across occupations using images specific to each race, which we provide in Figure~\ref{fig:maxskew-by-race}. All VLMs exhibit variation in gender bias across different races, highlighting the importance of measuring bias in the presence of other social attributes. While CLIP has lower overall skewness than other evaluated models, it also has the greatest disparity in $\textrm{MaxSkew}@K$ across different races; specifically, CLIP exhibits nearly 2x more gender skewness for images depicting Indian subjects than it does for Latinos.  Interestingly, both CLIP and FLAVA exhibit the greatest gender bias in retrieving images depicting Indian people, while SLIP and ALIP exhibit the greatest gender bias for Middle Eastern people.

As a further case study in uncovering intersectional social biases in VLMs, Figure~\ref{fig:doctor} provides a breakdown of the proportion of images retrieved using gender- and race-neutral prompts for the `Doctor' occupation. The left and center plots of Figure~\ref{fig:doctor} depict the marginal distributions of retrievals for each race and gender (respectively), whereas the right plot provides the distribution of retrieved images for intersectional race-gender attributes. While SLIP, OpenCLIP, LaCLIP, and ALIP exhibit strong bias for retrieving male images, this gender bias occurs along starkly different dimensions with respect to race. OpenCLIP strongly favors retrieving images of Asian male doctors when presented with a neutral text prompt,
whereas the other three models prefer to retrieve images of Indian male doctors. This gender bias is inverted for CLIP, which retrieves images of women in a slightly higher proportion than men, but still has zero representation of certain groups (e.g., Black females). These results show that seemingly similar biases among pre-trained VLMs for social attributes such as gender can interact very differently with other attributes such as race, which highlights the importance of studying bias in the presence of intersectional social attributes.
\section{Mitigating Intersectional Biases}
\label{sec:mitigating}

\begin{table*}[h!]
\footnotesize
\begin{center}
% \resizebox{1\columnwidth}{!}
% {
\begin{tabular}{l c c c c c c} %
\toprule
& \multicolumn{2}{c}{\textbf{CLIP}~\cite{radford2021learning}} & \multicolumn{2}{c}{\textbf{ALIP}~\cite{yang2023alip}} & \multicolumn{2}{c}{\textbf{FLAVA} ~\cite{singh2022flava}}  \\
\cmidrule(lr){2-3}
\cmidrule(lr){4-5}
\cmidrule(lr){6-7}
\textbf{Intersectional Bias} & \textbf{Pre-trained} & \textbf{Debiased} & \textbf{Pre-trained} & \textbf{Debiased} & \textbf{Pre-trained} & \textbf{Debiased}\\
\midrule
$(\textrm{Race}, \textrm{Gender})$ & 1.02 & \textbf{0.77} & 1.40 & \textbf{1.16} & 0.98 & \textbf{0.79}\\
$(\textrm{Physical Char.}, \textrm{Gender})$ & 0.87 & \textbf{0.71} & 1.28 & \textbf{1.02} & 0.92 & \textbf{0.81}\\
$(\textrm{Physical Char.}, \textrm{Race})$ & 1.19 & \textbf{0.87} & 1.55 & \textbf{1.13} & 1.12 & \textbf{0.97}\\
\bottomrule
\end{tabular}
% }
\caption{Mean of $\textrm{MaxSkew}@K$ for pre-trained and debiased variants of CLIP, ALIP, and FLAVA, estimated by withholding counterfactual sets for 20\% of the occupations in our dataset. Best results are in bold.}
\label{tab:intersectional-debias}
\end{center}
\end{table*}

We investigate the suitability of our SocialCounterfactuals dataset for debiasing VLMs through additional training. \anahita{Recommend rewording this}

\subsection{Training Experiment Setting}

For each of the three segments of our dataset (see Table~\ref{tab:dataset-details}), we withhold counterfactual sets associated with 20\% of the occupation subjects for testing and use the remainder as a training dataset. We then separately finetune ALIP, CLIP, and FLAVA on each of these three training datasets, which we hereafter refer to as the `debiased' variants of these models. To estimate the magnitude of debiasing, we evaluate each model's $\textrm{MaxSkew}@K$ for intersectional bias using the withheld testing dataset containing 20\% of the occupations. 

\subsection{Results for Intersectional Social Biases}

Table~\ref{tab:intersectional-debias} provides the mean $\textrm{MaxSkew}@K$ calculated over our withheld test sets for pre-trained and debiased variants of CLIP, ALIP, and FLAVA. 
Training on our dataset has the greatest overall debiasing effect on ALIP, producing up to a 0.426 absolute reduction in $\textrm{MaxSkew}@K$.
Among the three segments of our dataset used for debiasing, $\textrm{MaxSkew}@K$ for $(\textrm{Physical Char.}, \textrm{Race})$ intersectional bias has the greatest amount of skewness in pre-trained models. 
However, training on our dataset also produces the greatest absolute reduction in skewness for this type of intersectional bias, with CLIP having an absolute reduction in $\textrm{MaxSkew}@K$ of 0.327 for $(\textrm{Physical Char.}, \textrm{Race})$ intersectional biases. 

These results show that training VLMs with our dataset produces significant debiasing effects across all three types of intersectional biases.
Furthermore, the strongest intersectional biases observed in pre-trained VLMs benefit the most from debiasing.\anahita{This sentence is too verbose}\phillip{I split it into two sentences. Please check it again}
Remarkably, these debiasing effects are observed despite there being no overlap between the occupations used for training and testing, which demonstrates that the debiasing effects generalize to new subjects not seen during training.
This suggests that the ability of our dataset to mitigate intersectional bias in VLMs is not limited to only the occupation subjects that we investigated.

\subsection{Analysis of Race-Gender Debiased CLIP Model}

To further understand the impact of training VLMs using our synthetic counterfactuals, we conduct analyses of CLIP after debiasing for Race-Gender intersectional bias. For simplicity, we hereafter refer to this model as Debiased CLIP.

\paragraph{Skewness evaluation using other datasets with real image-text pairs.} Since our dataset was synthetically generated, a natural question to ask is how well our observed debiasing effects extend to evaluations with real image-text pairs.
Unfortunately, there are no such existing resources for measuring the intersectional social biases that we investigate in this work.
However, several real image-text datasets have been proposed for evaluating (marginal) social biases for attributes such as perceived race and gender.

To evaluate our Debiased CLIP model on such datasets, we use the Protected-Attribute Tag Association (PATA) dataset introduced in \cite{seth2023dear} for nuanced reporting of biases associated with race, age, and gender protected attributes. The PATA dataset comprises of 4,934 public images organized in 24 scenes, where the scenes represent situations in which certain groups of humans may have biases associated with them. The images are annotated with binary gender (male, female), ethnic-race labels (White, Black, Indian, East Asian, Latino-Hispanic) and two age groups (young and old). We also evaluate our Debiased CLIP model on the VisoGender dataset \cite{hall2023visogender}, which was curated to benchmark gender bias in image-text pronoun resolution. VisoGender consists of 690 images of people in 23 unique occupational settings.\footnote{We compare SocialCounterfactuals to PATA and VisoGender using the FID and IS metrics in Section~\ref{app:fid-and-is}.}

Table~\ref{tab:other-bias-datasets} provides the mean $\textrm{MaxSkew}@K$ of our Debiased CLIP model on these two datasets. Despite only being trained for mitigating intersectional bias using synthetic examples, our Debiased CLIP model achieves a 15\% and 12\% relative reductions in skewness when measured using real image-text examples from VisoGender and PATA (respectively). Both models have much lower $\textrm{MaxSkew}@K$ for these datasets than for SocialCounterfactuals, which demonstrates how our dataset reveals significantly more skewness in retrieval results than existing single-attribute datasets.

\begin{comment}
    \begin{table}
\footnotesize
\begin{center}
% \resizebox{1\columnwidth}{!}
% {
\begin{tabular}{l c c} %
\toprule
\textbf{Model} & \textbf{VisoGender}  & \textbf{PATA}\\
\midrule
Pre-trained CLIP & 0.27 & 0.16\\
Debiased CLIP & \textbf{0.23} & \textbf{0.04}\\
\bottomrule
\end{tabular}
% }
\caption{$\textrm{MaxSkew}@K$ of our debiased CLIP model as well as pre-trained CLIP on the VisoGender ~\cite{hall2023visogender} and PATA ~\cite{seth2023dear} datasets}
\label{tab:other-bias-datasets}
\end{center}
\end{table}
\end{comment}

 \begin{table}
\footnotesize
\begin{center}
% \resizebox{1\columnwidth}{!}
% {
\begin{tabular}{l c c} %
\toprule
\textbf{Model} & \textbf{VisoGender} \cite{hall2023visogender} & \textbf{PATA} ~\cite{seth2023dear}\\
\midrule
Pre-trained CLIP & 0.269  & 0.323\\
Debiased CLIP & \textbf{0.219} & \textbf{0.283}\\
\bottomrule
\end{tabular}
% }
\caption{$\textrm{MaxSkew}@K$ of our debiased CLIP model as well as pre-trained CLIP on the VisoGender ~\cite{hall2023visogender} and PATA ~\cite{seth2023dear} datasets.}
\label{tab:other-bias-datasets}
\end{center}
\end{table}

\paragraph{Impact of debiasing on task-specific performance.} An important question for practitioners seeking to debias VLMs is the extent to which debiasing degrades the performance of the model on other tasks. As described previously in Section~\ref{sec:intro}, social biases can be viewed as a type of spurious correlation which models learn as shortcuts to achieving high performance on training datasets. Consequently, it is expected that eliminating these shortcuts may degrade the performance of the model on other tasks to some extent.

We estimate the magnitude of this impact by evaluating our Debiased CLIP model on image-text retrieval and zero-shot image recognition tasks. Table~\ref{tab:zero-shot-itr} provides the text retrieval and image retrieval performance of both pre-trained CLIP and our Debiased CLIP on Flickr30K. We observe that our Debiased CLIP model achieves equivalent or better performance across all three text retrieval settings compared to pre-trained CLIP. In the image retrieval settings, our Debiased CLIP model exhibits a minor degradation in performance compared to pre-trained CLIP.

\begin{table}
\footnotesize
\begin{center}
\resizebox{1\columnwidth}{!}
{
\begin{tabular}{l c c c c c c} %
\toprule
& \multicolumn{3}{c}{\textbf{Text Retrieval}} & \multicolumn{3}{c}{\textbf{Image Retrieval}}  \\
\cmidrule(lr){2-4}
\cmidrule(lr){5-7}
\textbf{CLIP Model} & \textbf{R@1} & \textbf{R@5} & \textbf{R@10} & \textbf{R@1} & \textbf{R@5} & \textbf{R@10} \\
\midrule
Pre-trained & 67.1 & 89 & 93.8 & \textbf{69.4} & \textbf{90.6} & \textbf{94.9}\\
Debiased & \textbf{69.2} & \textbf{89.6} & 93.8 & 67.3 & 90.4 & 93.8\\
\bottomrule
\end{tabular}
}
\caption{Image \& text retrieval performance of Debiased CLIP and pre-trained CLIP on Flickr30K\cite{young2014image}. Debiasing CLIP with our dataset results in improved performance on text retrieval and only minimal performance degradation on image retrieval.}
\label{tab:zero-shot-itr}
\end{center}
\end{table}

Table~\ref{tab:zero-shot-image-recognition} measures the accuracy of pre-trained CLIP and our Debiased CLIP model on zero-shot image recognition datasets. Similar to the image retrieval evaluation, our Debiased CLIP model exhibits a minor degradation in accuracy on these datasets. These performance reductions are similar in magnitude to those observed in prior work on debiasing CLIP \citep{seth2023dear} and characterize the trade-off between model fairness and absolute performance inherent to debiasing efforts. 

Depending on the intended use case, the benefits of reducing skewness in the retrieval results of VLMs may be far more valuable than the relatively small decrease in performance observed in debiased models. Additionally, these results were obtained without any tuning of the training process for balancing task-specific performance with debiasing efforts. We hypothesize that additional attention to these task-specific performance measures during training, as well as other strategies such as mixing real data with our counterfactual examples, may produce a different balance between model debiasing and task-specific performance.

\begin{table}
\footnotesize
\begin{center}
\resizebox{1\columnwidth}{!}
{
\begin{tabular}{l c c c c} %
\toprule
% \textbf{CLIP Model} & \textbf{CIFAR10}\cite{krizhevsky2009learning} & \textbf{CIFAR100}\cite{krizhevsky2009learning} & \textbf{Caltech101}\cite{fei2004learning} & \textbf{Caltech256}\cite{griffin2007caltech} & \textbf{ImageNet}\cite{5206848} \\
\textbf{CLIP Model} & \textbf{CIFAR10}\cite{krizhevsky2009learning} & \textbf{CIFAR100}\cite{krizhevsky2009learning} & \textbf{Caltech256}\cite{griffin2007caltech} & \textbf{ImageNet}\cite{5206848} \\
\midrule
% Pre-trained & \textbf{88.8} & \textbf{64.17} & \textbf{90.32} & \textbf{83.43} & \textbf{59.25} \\
% Debiased &  86.72 & 61.46 & 84.38 & 79.72 & 55.38\\
Pre-trained & \textbf{88.80} & \textbf{64.17} & \textbf{83.43} & \textbf{59.25} \\
Debiased &  86.72 & 61.46 & 79.72 & 55.38\\
\bottomrule
\end{tabular}
}
\caption{Accuracy of our debiased CLIP model as well as pre-trained CLIP on zero-shot image recognition datasets. Debiasing CLIP with our dataset results in minimal performance degradation.\phillip{Should we drop Caltech101 here? I'm not sure that we need six datasets and it would make the table easier to read} \anahita{Maybe also keep only one CIFAR dataset?}}\phillip{Do we need to add citations for these datasets?}
\label{tab:zero-shot-image-recognition}
\end{center}
\end{table}

\phillip{Examples of debiasing impact on occupation-specific retrieval skewness.}
\section{Conclusion}
\label{sec:conclusion}

In this work, we presented a methodology for automatically generating counterfactual examples for probing and mitigating intersectional bias in VLMs. 
We used our approach to construct SocialCounterfactuals, a large dataset of image-text counterfactuals depicting intersectional social attributes related to gender, race, and physical characteristics for various occupations. 
Our evaluations of six pre-trained VLMs showed that all exhibit significant intersectional social biases in retrieval results, with substantial variation in retrieval skewness across differing racial and gender attributes.
Through training experiments, we further demonstrated that SocialCounterfactuals can be a valuable resource for mitigating skewness in VLMs.
A promising direction for future work could be extending our approach to investigate intersectional social biases in VLMs for other attributes and subjects.
Our SocialCounterfactuals dataset could also be a valuable resource for reducing bias in generative text-to-image diffusion models.
Finally, alternative training strategies for debiasing VLMs with synthetic counterfactuals could be explored to balance bias mitigation with task-specific performance measures. 
\paragraph{Limitations and Ethical Considerations}
\label{sec:limitations}

Despite our best efforts, the templates and methodologies we adopt may themselves contain some latent biases which could contribute to the implicit biases exhibited by VLMs. All statements pertaining to gender, race, and occupational attributes or associations should be interpreted only within the context of our experiments. Furthermore, all discussion of social attributes in this work are intended to be interpreted as \textit{perceived}. We are aware that our approach only considers binary classification of genders and does not exhaustively encompass all races, physical characteristics, and occupations, which is due to data limitation rather than our value judgements. 
% The results we present cannot be assumed to generalize to social and demographic terms omitted in our analysis. The labels for the attributes we present in the paper are derived from prior work and were further limited to those which stable diffusion could depict. Our goal is to provide text labels that produce perceived physical differences, but these are not labels we aim to impose on any groups or sub-groups. Similar to \citet{smith2022m}, we recognize there are trade-offs in creating lists of socials attributes. While these lists may not be entirely inclusive, we leverage them for their benefit in identifying and mitigating bias. 
% Any absence of groups or sub-groups should not be considered as disregard or our own bias, but to be treated as a data contraint.
% Our study was conducted in English, which limits the generalizability of our findings to other languages. 

With the findings we present in this paper, we aim to increase the understanding of bias in VLMs and probe mitigation strategies. We acknowledge that our work does not encompass all possible social attributes and that our selected categories for gender, race, physical characteristics, and occupations may harbor stereotypes that cannot be assumed to represent their entire groups. 
% Similar to  \citet{hall2023visogender}, we recognize that we may miss intersectional characteristics that constitute a well-accepted image of a person in a specific occupation or belonging to a race. 
Our aim is that the techniques presented in this work can help reduce various social disparities in VLMs and can be further extended to include more genders, races, occupations and other social characteristics. 
Continuing these efforts will increase confidence in the ability of VLMs to exhibit fairness with respect to differing social attributes. See Section~\ref{sec:additional-limitations} for additional discussion.
% We understand that the use of a bias reduction strategy without deep understanding of various nuances does not guarantee a foolproof solution in bias elimination, and still may result in VLMs that cause harm and stigmatize certain subsets of individuals. Therefore, debiasing efforts should be further developed prior to wide-spread adoption in sensitive applications.
%\input{sections/8_ethical}

{
    \small
    \bibliographystyle{ieeenat_fullname}
    \bibliography{main}
}

\clearpage
\setcounter{page}{1}
\maketitlesupplementary

% \newpage

% \appendix
% \section{Appendix}
\begin{comment}
    \begin{table*}
\footnotesize
\begin{center}
% \resizebox{1\columnwidth}{!}
% {
\begin{tabular}{l c c c c c c} %
\toprule
& \multicolumn{2}{c}{\textbf{CLIP}~\cite{radford2021learning}} & \multicolumn{2}{c}{\textbf{ALIP}~\cite{yang2023alip}} & \multicolumn{2}{c}{\textbf{FLAVA} ~\cite{singh2022flava}}  \\
\cmidrule(lr){2-3}
\cmidrule(lr){4-5}
\cmidrule(lr){6-7}
\textbf{Intersectional Bias} & \textbf{Pre-trained} & \textbf{Debiased} & \textbf{Pre-trained} & \textbf{Debiased} & \textbf{Pre-trained} & \textbf{Debiased}\\
\midrule
$(\textrm{Race}, \textrm{Gender})$ & 0.20 & 0.19 & 0.24 & 0.23 & 0.21 & 0.20\\
$(\textrm{Physical Char.}, \textrm{Gender})$ & 0.20 & 0.19 & 0.23 & 0.21 & 0.20 & 0.19\\
$(\textrm{Physical Char.}, \textrm{Race})$ & 0.20 & 0.19 & 0.20 & 0.19 & 0.18 & 0.17\\
\bottomrule
\end{tabular}
% }
\caption{NDKL scores for pre-trained and debiased variants of CLIP, ALIP, and Flava, estimated by withholding counterfactual sets for 20\% of the occupations in our dataset.}
\label{tab:intersectional-debias-ndkl}
\end{center}
\end{table*}
\end{comment}

\section{Additional Analysis of SocialCounterfactuals}
\subsection{Examples of Counterfactual Sets}
\label{app:additional-examples}

\begin{figure*}[ht!]
    \centering
    \includegraphics[trim={2mm 2mm 2mm 
    2mm},clip,height=1\textheight]{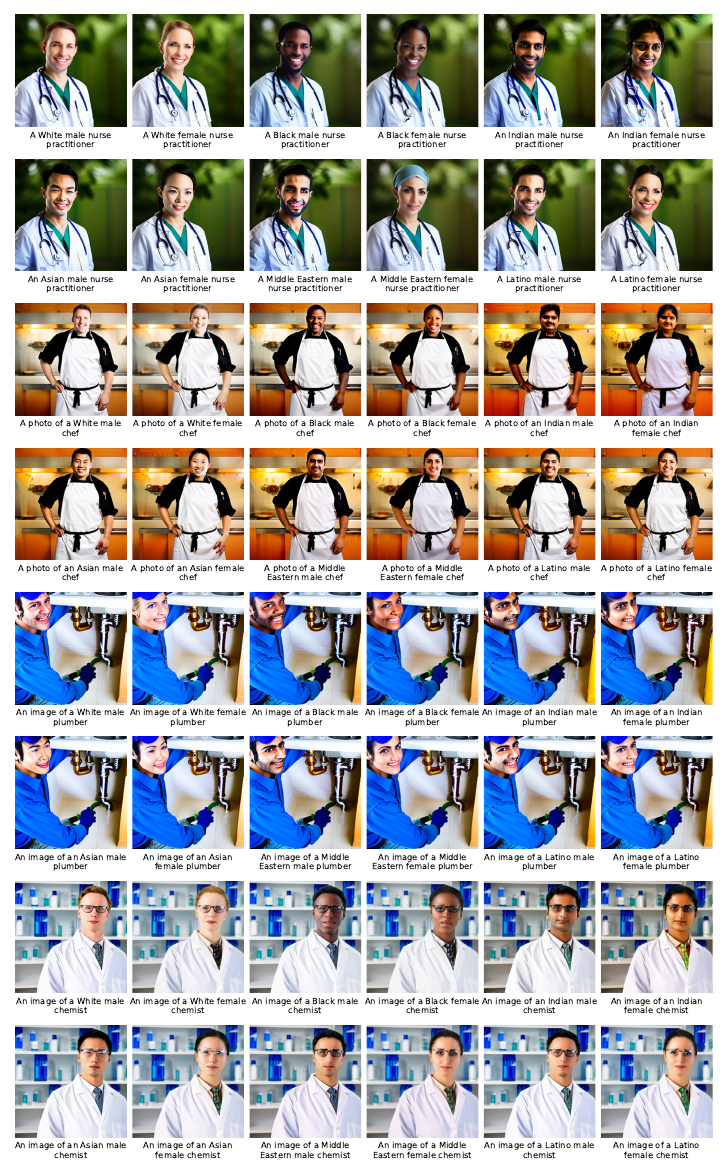}
    \caption{Additional examples of counterfactual sets produced by our approach for the (Race, Gender) attribute pair.}
    \label{fig:occupation-race-gender-examples}
\end{figure*}

\begin{figure*}[ht!]
    \centering
    \includegraphics[trim={2mm 2mm 2mm 
    2mm},clip,width=1\textwidth]{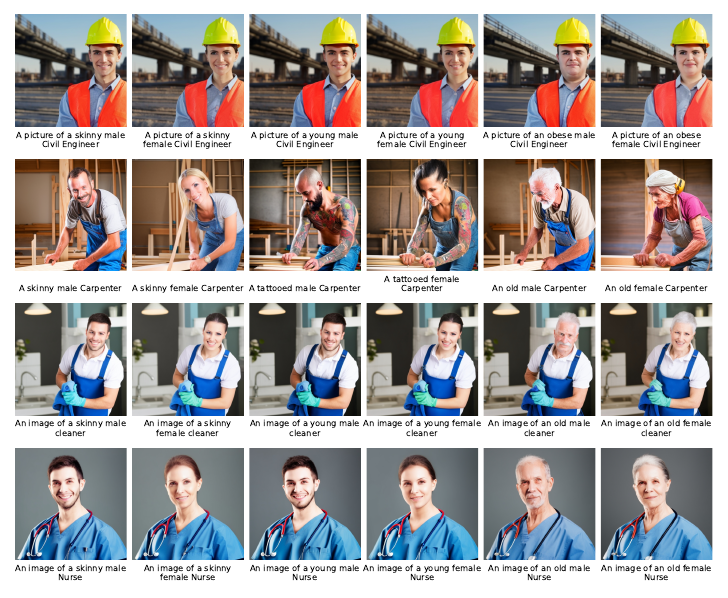}
    \caption{Additional examples of counterfactual sets produced by our approach for the (Physical Characteristics, Gender) attribute pair.}
    \label{fig:occupation-physical-gender-examples}
\end{figure*}

\begin{figure*}[ht!]
    \centering
    \includegraphics[trim={2mm 2mm 2mm 
    2mm},clip,width=1\textwidth]{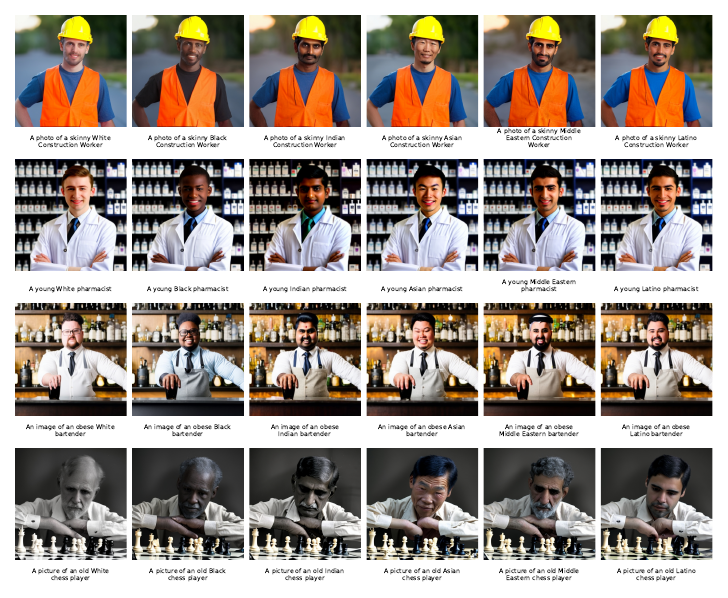}
    \caption{Additional examples of counterfactual sets produced by our approach for the (Physical Characteristics, Race) attribute pair.}
    \label{fig:occupation-physical-race-examples}
\end{figure*}

Figure~\ref{fig:occupation-race-gender-examples}, 
~\ref{fig:occupation-physical-gender-examples} and 
~\ref{fig:occupation-physical-race-examples} provides additional examples of counterfactual sets generated by our approach.

\subsection{Error Analysis}
\label{sec:error-analysis}
To investigate the different failure modes for the generated images in this study, we conducted a human evaluation of counterfactual sets for gender-race bias in occupations. We sampled 1200 images (100 for each gender-race combination) prior to our third stage of filtering (CLIP Attribute Detectability Filtering) and then annotated them into 5 categories. Results are shown in Table \ref{table:error-analysis}. We found that 90.8\% of the images were correctly generated in terms of occupation, gender and race. In 5.2\% of the samples, the gender could not be identified. This was typically due to subjects looking backwards or facing the wrong direction. 

\textit{Failure to generate female subjects} was the second most frequent error with 2.2\%, followed by \textit{subject completely out of frame/focus}. The least common error was \textit{Failure to generate male subject} with a frequency of only 0.8\%.  No failures related to race were observed.
% It seems that the model uses the most stereotypical features for each of the races and consequently race mismatches would be very hard to prove. 
Sampled images illustrating each of the different modes of failures are displayed in Table \ref{table:error-examples}.

\begin{table}[h!]
 \centering
 \resizebox{1\columnwidth}{!}
{
 \begin{tabular}{l c c} 
 \hline
 \textbf{Error Category} & \textbf{\% present in sample} \\ [0.5ex]
 \hline
 Good & 90.8\% \\ 
 Cannot discern gender & 5.2\% \\
 Failure to generate female subject & 2.2\% \\
 Subject completely out of frame/focus & 1.0\% \\
 Failure to generate male subject & 0.8\% \\ [1ex] 
 \hline
\end{tabular}
}
\vspace{1mm}
\caption{Error analysis for 1200 random samples focused on gender and race.}
\label{table:error-analysis}
\end{table}
\begin{table*}
\centering
% \small
\resizebox{1\textwidth}{!}
{%
\begin{tabular}{p{0.1cm} p{6.8cm} p{0.1cm} p{6.8cm}}

\midrule

\rotatebox[origin=c]{90}{Failure to generate female subject/object} & 
\begin{minipage}{0.4\textwidth} 
    \centering 
    \includegraphics[width=1\textwidth]{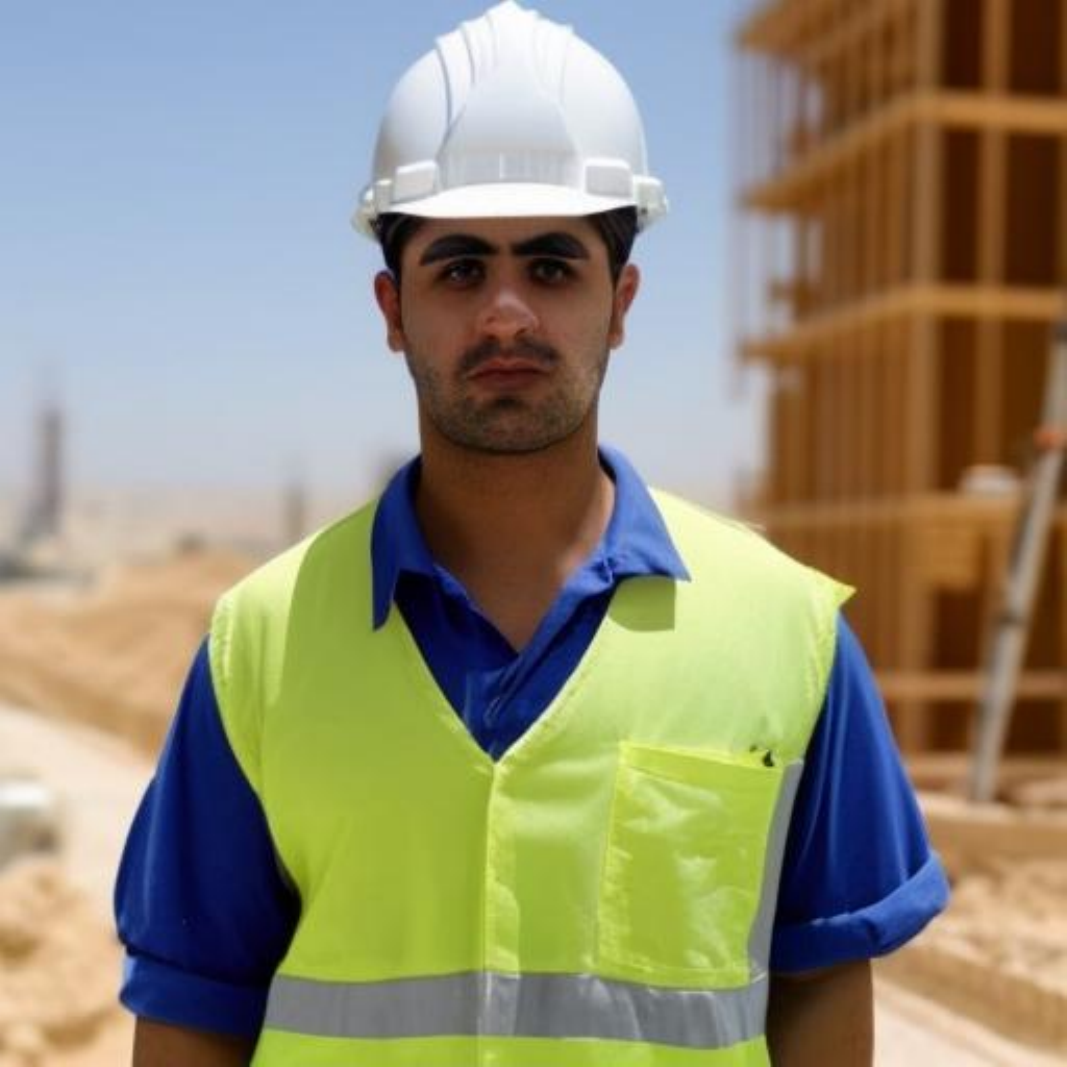} 
\end{minipage} & 
\rotatebox[origin=c]{90}{Cannot discern gender} & 
\begin{minipage}{0.4\textwidth} 
    \centering
    \includegraphics[width=1\textwidth]{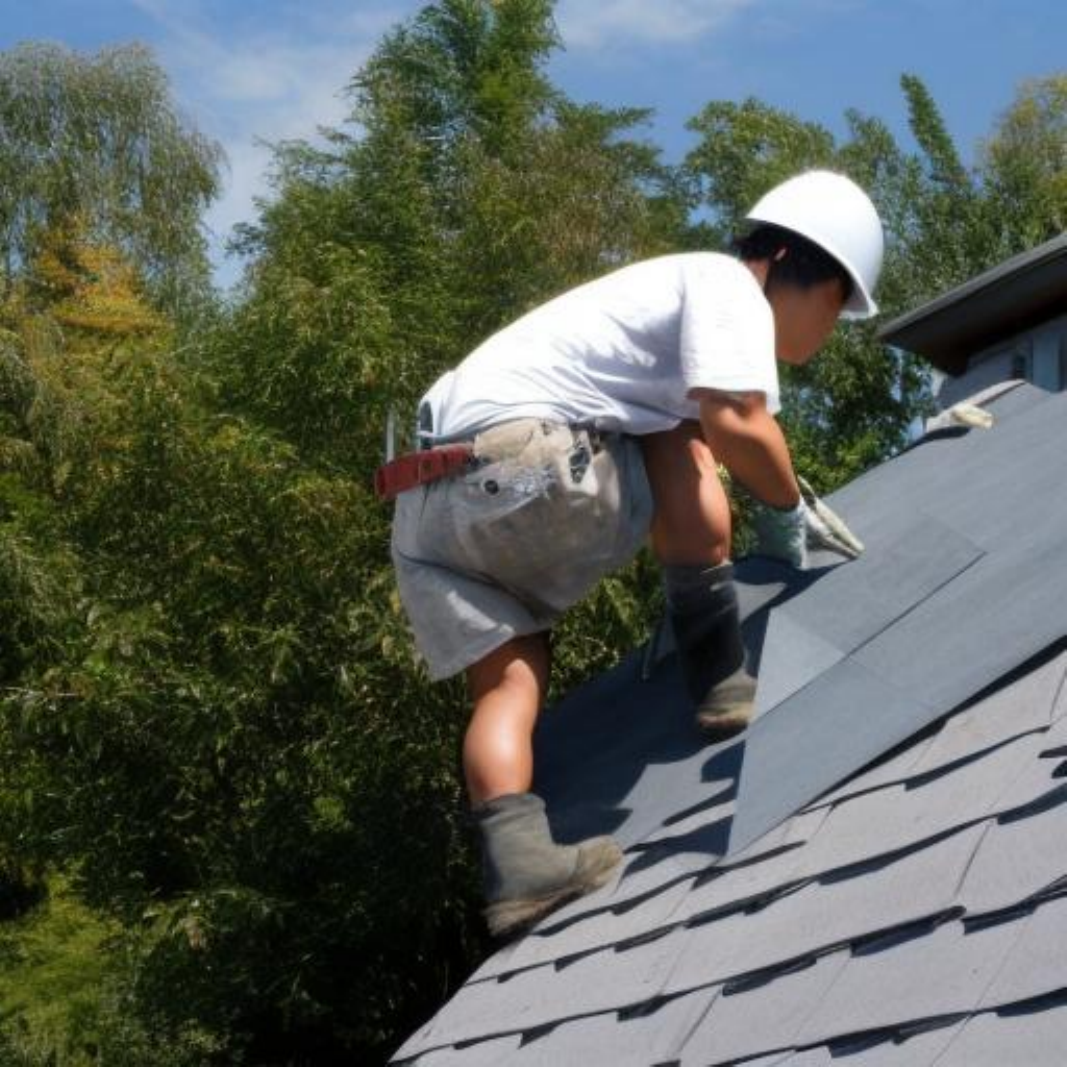}
\end{minipage} 
\vspace{0.1cm}\\ & 
\multicolumn{1}{c}{\it{a Middle Eastern female construction worker}} & &
\multicolumn{1}{c}{\it{a picture of an Asian female roofer.}} \\

\midrule

\rotatebox[origin=c]{90}{Failure to generate male subject/object} & 
\begin{minipage}{0.4\textwidth} 
    \centering 
    \includegraphics[width=1\textwidth]{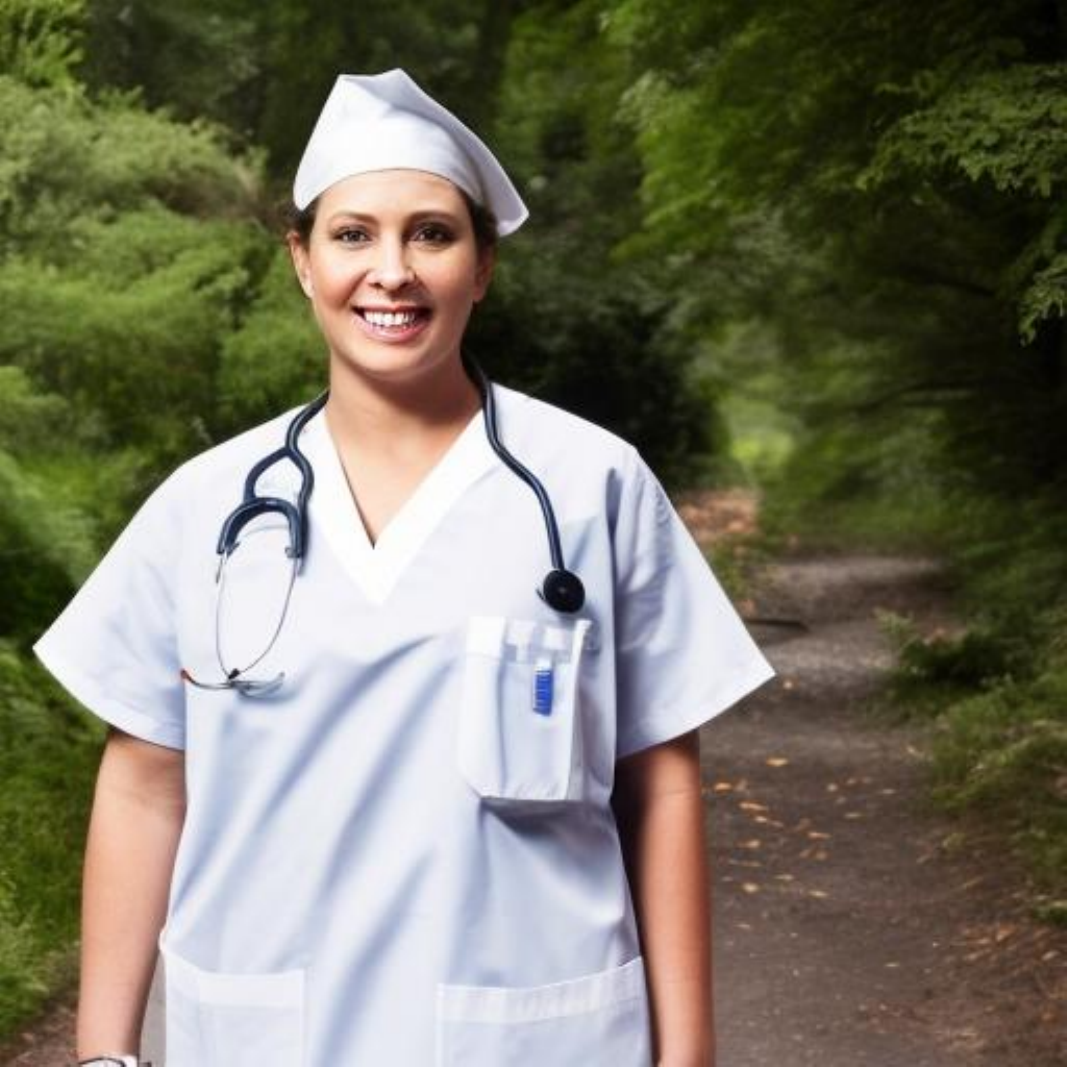} 
\end{minipage} &
\rotatebox[origin=c]{90}{Subject completely out of frame/focus} & 
\begin{minipage}{0.4\textwidth} 
    \centering
    \includegraphics[width=1\textwidth]{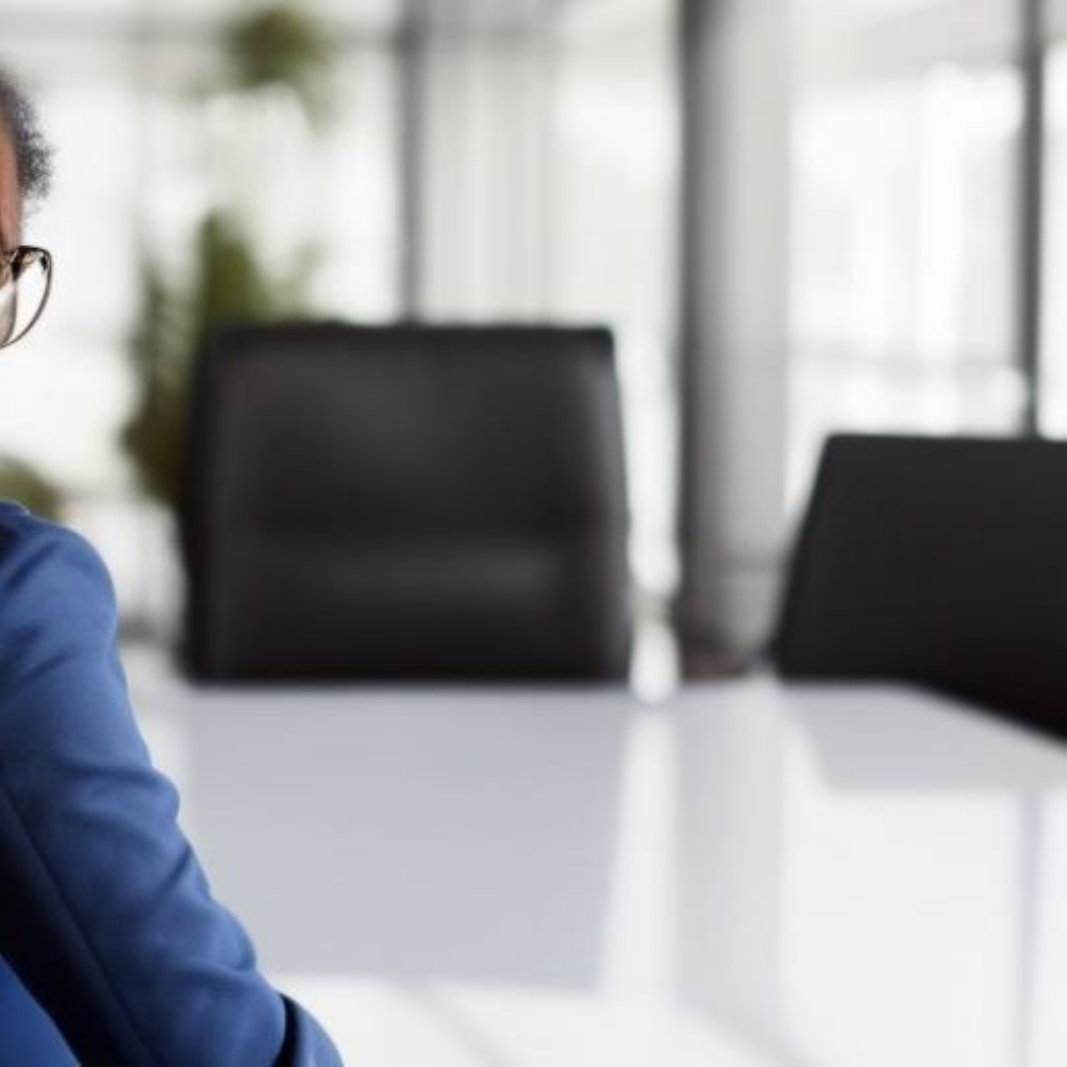} 
\end{minipage} 
\vspace{0.1cm}\\ & 
\multicolumn{1}{c}{\it{a White male nurse}} & &
\multicolumn{1}{c}{\textit{an image of a Black female chief executive officer}} \\

\bottomrule

\end{tabular}}
\caption{Examples of failure cases identified by manual error analysis}
\label{table:error-examples}
\end{table*}

An extended breakdown by race is shown in Table \ref{table:error-analysis-race}. We observe that across White, Black, Indian and Asian races, failure to generate either female or male subjects is approximately the same. For Latino images, there is a higher proportion of failures to generate female subjects; for Middle Eastern (M.E.) subjects, the proportion is even more pronounced, with a 12\% failure to generate female subjects. In contrast, no generation failures were observed for Middle Eastern male subjects. 

To further study the impact of our filtering method on image quality, we measure the same generation failure percentages after applying CLIP Attribute Detectability Filtering. This final stage of filtering increases the percentage of \textit{Good} images to 99\% and 96\% for male and female (respectively). Since the proportion of male and female are the same, the overall accuracy increases from 90.8\% to 97.5\% when CLIP Attribute Detectability Filtering is used.

\begin{table*}[h!]
 \centering
 \begin{tabular}{l c c c c c c} 
 \hline
 \textbf{Error Category} & 
 \textbf{White} & 
 \textbf{Black} & 
 \textbf{Indian} & 
 \textbf{Asian} & 
 \textbf{M.E.} & 
 \textbf{Latino} \\ [0.5ex]
 \hline
 Failure to generate female subject & 3\%  & 4\%  & 1\%  & 2\%  & 12\%  & 5\% \\ 
 Failure to generate male subject   & 4\%  & 2\%  & 1\%  & 1\%  & 0\%  & 2\% \\ [1ex] 
 \hline
\end{tabular}
\vspace{1mm}
\caption{Error analysis with 100 samples for each race-gender combination}
\label{table:error-analysis-race}
\end{table*}

% \subsection{Gender Prediction with CLIP}
% To investigate methods for automatically detecting incorrectly generated images, we used CLIP for text-to-image retrieval to generate similarity scores. Specifically, similarity scores were obtained for queries in the form of \textit{A male person} and \textit{A female person} against the set of generated images. If the score for \textit{A male person} was highest, then we can infer that the gender of the person generated is male, and vice versa. We compared these predictions against the human annotated samples detailed previously. In general, CLIP accurately identifies male and female subjects (See Table \ref{table:confusion-matrix}). Precision, recall and f1-scores are very high for classifying both male and female subjects. Therefore, it is our conclusion that CLIP could be an out-the-box tool which could be used to automate the detection of faulty images in terms of gender, thereby mitigating the need for human annotation. 

\begin{table}[h!]
 \centering
 \begin{tabular}{l c c c c c} 
 \hline
 \textbf{gender} & 
 \textbf{precision} &
 \textbf{recall} &
 \textbf{f1-score} &
 \textbf{support} \\ [0.5ex]
 \hline
 male & 0.99 & 0.95 & 0.97 & 549 \\ 
 female & 0.96 & 0.99 & 0.97 & 589 \\ [1ex] 
 \hline
\end{tabular}
\vspace{1mm}
\caption{Confusion Matrix for male and female detection.}
\label{table:confusion-matrix}
\end{table}

\subsection{Analysis of Occupation Detectability}
We manually evaluated images depicting all of the occupations listed in Table~\ref{tab:occupations} to identify those which are hard to recognize even for humans. In randomly selected subsets, we observed that occupations such as accountant, salesperson, economist, broker, web developer, attorney, banker, mathematician, and engineer produce very similar images depicting subjects wearing formal attire. Although the images accurately depict the social attributes (e.g., race, gender), the subject can be difficult to distinguish from other similar occupations. Because we evaluate skewness separately for each occupation, this similarity across occupations should not have a significant impact on the validity of bias estimation.
%Hence, for our experiments we use this curated list as specified in Table \anahita{Phillip/Tiep do we want a table with shortlisted occupations? If not, we can remove this sentence.}

We also observed that it can be difficult to distinguish subjects across the various the medical profession, (e.g., pediatrician, audiologist) since they are depicted by Stable Diffusion as generic doctors wearing coats or scrubs. Similar depictions were also seen for education-related professions (e.g., teacher, special ed teacher, primary school teacher). Images for the `pensioner' profession typically depict elderly individuals, which might be typical of the occupation but not necessarily sufficient to determine it from an image. We also observed that images generated of football player depict American football for White, Black and Latino races, while American soccer is depicted for this occupation for Indian, Asian and Middle Eastern races. We hypothesize that this is due to the fact that American soccer is often referred to as football in countries which are predominately populated by the latter three races, which influences how Stable Diffusion depicts this occupation for those races. 

The depiction of some occupations were observed to contain intrinsic biases related to physical characteristics. For example, a majority of the images analyzed for the computer programmer and software developer professions depicted individuals wearing glasses. For umpires, security guards, opera singers, we observed that most of the images depicted individuals who were perceived to be overweight or obese. Images generated for farmers were perceived as having darker skin tones across all races, indicating the greater sun exposure for individuals with occupation.  While not representative of all possible biases, these observations indicate the presence of some intrinsic bias associated with certain physical characteristics and occupations. 

Similarly, we observed certain gender biases associated with occupations. For example, Stable Diffusion failed to generate male subjects across all races for the midwife and maid professions in the random subset of examples we analyzed. This could could be due to how these occupations are closely associated with the female gender, which makes it challenging for Stable Diffusion to generate male counterfactual images. However, no such bias was observed while generating images of policeman or handyman for female subjects across various races. We observed some shortcomings in the depictions of Middle Eastern and Latino female priests, possibly indicating a strong association between this occupation and the male gender for certain races.

Finally, for images which depicted multiple individuals, we occasionally observed a mix-up in the race associated with the occupation. For instance, images depicting hairdressers and barbers often associated the described race with the client rather than the hairdresser or barber. This is likely due to challenges that Stable Diffusion has with accurately binding attributes when multiple subjects are present in an image.
% We also observed a mix-up in the race for hairdressers and barbers wherein the model seems to have mixed up the race of the client and person with the occupation.

\subsection{Generating counterfactuals for other subjects and social attributes}
\label{app:other-subjects-and-attributes}

In addition to generating counterfactual sets depicting occupations, we also explored the viability of using personality traits as the subject in our captions. Specifically, we used the same list of 63 personality traits as in \citet{naik2023social} to construct captions in a similar manner as described in Section~\ref{sec:captions}, using the template \textrm{<$p$> <$s$> <$a_{1}$> <$a_{2}$>}. For example, given the personality trait \textit{helpful}, we constructed captions for investigating race-gender bias such as \textit{A helpful white man}, \textit{A helpful black woman}, etc. However, we determined after a manual evaluation of generated images that Stable Diffusion struggled to depict these personality traits, which in turn degraded the quality of depictions of investigated social attributes. We therefore decided to omit images for these types of subjects from SocialCounterfactuals.

Beyond the race, gender, and physical characteristic social attributes, we also investigated the use of religion as a fourth type of social attribute for our counterfactual sets. This social attribute set included the terms \textit{Christian, Muslim, Hindu, Buddhist, Atheist}, and \textit{Agnostic}. After manual evaluation of images generated using religion as a social attribute, we determined that several of the religion terms consistently produced nearly identical images (e.g., \textit{Christian, Atheist, Agnostic}). Among images that produced discernible differences, Stable Diffusion primarily used race to differentiate subjects for each religion. Therefore, we decided to exclude counterfactual sets involving the religion social attribute from SocialCounterfactuals, and leave further investigation into intersectional biases involving religion to future work.

\section{Dataset Generation \& Experiment Details}

\subsection{Details of compute infrastructure used}
The counterfactual image-text data was created using a large AI cluster equipped with Intel 3\textsuperscript{rd} Generation Intel\textsuperscript{®} Xeon\textsuperscript{®} processors and Intel\textsuperscript{®} Gaudi-2\textsuperscript{®} AI accelerators. Up to 256 Intel Gaudi-2\textsuperscript{®} AI accelerators were used to generate our SocialCounterfactuals dataset. 

\subsection{Details of training experiments}
\label{app:training-details}
We fine-tune CLIP with learning rates of 7e-6, 5e-6 and 9e-6 for (Race, Gender), (Physical Char., Gender) and (Physical Char., Race) respectively for 1 epoch and a batch size of 32. For the FLAVA and ALIP, we follow the same fine-tuning setting as CLIP.

\subsection{Details of dataset filtering}
\label{app:dataset-filtering-detail}
Table \ref{table:filtering-stats} provides details of the quantity of counterfactual sets remaining after each stage of filtering with our methodology. The total number of original counterfactual sets was 312,000, which corresponds to 5,408,000 generated images. After applying the CLIP similarity filter, the dataset was reduced to 21,359 counterfactual sets. This stage of filtering removes a significant portion of our generated dataset, but helps ensure that only images with the highest quality are retained. Once the NSFW filter was applied, the number of counterfactual sets decreased to 21,116, representing a 1\% reduction. Finally, after the CLIP attribute detectability filter was applied, the dataset was reduced to 13,824 counterfactual sets, which is the final size of SocialCounterfactuals reported in Table \ref{tab:dataset-details}. Below we provide additional details of each of the filtering stages.

\paragraph{NSFW Filtering.}
A manual analysis of the generated images detected NSFW samples. In order to detect and discard these images, we used an off-the-shelf fine-tuned vision transformer (google/vit-base-patch16-224-in21k) trained for NSFW image classification\footnote{\url{https://huggingface.co/Falconsai/nsfw_image_detection}}. This model was fine-tuned used a proprietary dataset containing 80,000 images which was carefully curated to represent two classes: \textit{nsfw} and \textit{normal}. The reported evaluation accuracy of the fine-tuned model is 98.03.

After applying the NSFW classification model to our dataset, approximately 0.9\% of the counterfactual sets were discarded for the (Race, Gender) segment, 1.4\% for (Physical Char., Gender) segment, and 2.7\% for (Physical Char., Race) segment. See Table \ref{table:filtering-stats} for details.

\paragraph{CLIP Attribute Detectability Filtering.}
To further ensure the quality of our dataset, we additionally filter counterfactual sets based on the ability of CLIP to detect and discern the targeted social attributes in each image. 
Intuitively, a counterfactual set is filtered out with respect to an attribute type if the number of its images that lack of detectability of that attribute type is less than a learnt threshold. Such a threshold is learnt based on the manual labels of 100 randomly sampled respective counterfactual sets. Then a counterfactual set is eventually filtered out if it is filtered out with respect to some of its attributes.

In more detail, we employed a two-phase approach for CLIP attribute detectability filtering. Without the loss of generality, considering an arbitrary intersectional bias $(X, Y)$ and a target attribute type X. Initially, we randomly sampled 100 counterfactual sets from $(X, Y)$-segment of our dataset. In the first phase, we manual labeled whether each counterfactual set is filtered out based on how many of its images possess their attribute type $X$ discernibly. 

In the second phase, we develop a learnable threshold-based heuristic to filter out a counterfactual set if the number of its images that have their targeted attribute $X$ discernible by CLIP-based scores is less than the respective threshold. 
In particular, we use the names of all the attributes of the targeted attribute type  to construct the set of potential text pairings as \{ a/an $x$ person\} where $x$ is an attribute belonging to the attribute type $X$ and predict the most probable (image, text) pair according to CLIP-based image-text similarity scores. For instance, if $X$ is gender, then the potential text pairing set is \{a female person, a male person\}. An image is said to have their targeted attribute $x \in X$ discernible by CLIP-based score if the pair of the image and the text `a/an $x$ person' is the most probable one.
The learnable threshold were heuristically derived to obtain high correspondence between automatic filtering with CLIP-based scores and those filtered by the manual human annotation in the first phase. 
Table~\ref{tab:manual-clip-filtering-threshold} lists learnable thresholds for each attribute pair and its respective attributes. 

\begin{table}
\footnotesize
\begin{center}
\resizebox{1\columnwidth}{!}
{
\begin{tabular}{l c c c } %
\toprule
Attribute Pair & \multicolumn{3}{c}{Threshold}\\
\cmidrule(lr){2-4}
 &  Gender & Race & Physical Char.\\
 \midrule

$(\textrm{Race}, \textrm{Gender})$ & 12 & 9 & N/A \\
$(\textrm{Physical Char.}, \textrm{Gender})$ & 10 & N/A & 5 \\
$(\textrm{Physical Char.}, \textrm{Race})$ & N/A & 13 & 8\\
\bottomrule
\end{tabular}
}
\caption{CLIP attribute detectability filtering thresholds for each attribute type in each attribute pair. N/A means not applicable when an attribute type is not a part of the attribute pair.}
\label{tab:manual-clip-filtering-threshold}
\end{center}
\end{table}

We acknowledge that in spite our best efforts, it is possible this  manual analysis could propagate the annotator's bias associated with the various social attributes we investigate. While we also acknowledge that automatic filtering may introduce additional bias from CLIP, our error analysis in Section~\ref{sec:error-analysis} shows that this filtering increases the quality of our dataset (90.8\% $\rightarrow$ 97.5\%). Additionally, our quantitative results (Table~\ref{tab:intersectional-debias}) show significant debiasing of other models (e.g., FLAVA) after training on our dataset, even when measured using other non-synthetic datasets (Table~\ref{tab:other-bias-datasets}). This suggests that any additional bias introduced by our filtering method is less significant than the overall debiasing effect produced by training on our dataset.

Despite the impressive performance of text-to-image diffusion models, our error analysis (Section~\ref{sec:error-analysis}) shows they cannot be relied upon in an automated synthetic image generation pipeline without the use of filtering. Manual filtering by humans is not practical when generating a dataset at our scale (over 5.4 million images before filtering). Furthermore, counterfactual images depicting various combinations of intersectional social biases do not exist in natural image datasets. Therefore, we believe the use of automated filtering is necessary to construct a dataset which is useful for investigating intersectional social biases at scale. 
%

% At the end of the second phase, we achieved a threshold for each attribute type in each attribute type pair. A counterfactual set is filtered out due to an attribute type if the number of its constituent images whose corresponding targeted attribute is discernible by CLIP is less than the corresponding learn threshold. A counterfactual set is remained if it is not filtered out due to any related attribute type.

\begin{table*}[h!]
 \centering
 \resizebox{1\textwidth}{!} {
 \begin{tabular}{l c c c c c c c c c c} 
 \hline
 \thead{Attribute Types} & 
 \thead{No. Sets} & 
 \thead{Images} & 
 \thead{No.} & 
 \thead{Sets After} & 
 \thead{\%} & 
 \thead{Sets After} & 
 \thead{\%} & 
 \thead{Sets After} & 
 \thead{\%} & 
 \thead{Final No.} \vspace{-0.2cm} \\
 
  & 
  & 
 \thead{Per Set} & 
 \thead{Images} & 
 \thead{CLIP} & 
 \thead{Filtered} & 
 \thead{NSFW} & 
 \thead{Filtered} & 
 \thead{CLIP} & 
 \thead{Filtered} & 
 \thead{Images} \vspace{-0.2cm} \\
 
 & 
 & 
 & 
 & 
 \thead{Sim. Filter} & 
 \thead{Out} & 
 \thead{Filter} & 
 \thead{Out} & 
 \thead{Attrib. Filter} & 
 \thead{Out} & \\
 
 \hline
$(\textrm{Race, } \textrm{Gender})$ & 104,000 & 12 & 1,248,000 & 13,147 &	87\% & 13,035 & 0.9\% & 7,936 & 39\% & 95,232 \\ 
$(\textrm{Physical Char., } \textrm{Gender})$  & 104,000 & 10 & 1,040,000 & 7,254 & 93\% & 7,149 & 1.4\% & 5,052 & 29\% & 50,520 \\
$(\textrm{Physical Char., } \textrm{Race})$  & 104,000 & 30 & 3,120,000 & 958 & 99\% & 932 & 2.7\% & 836 & 10\% & 25,080 \\ [1ex] 
 \hline
& 312,000 &  & 5,408,000 & 21,359 && 21,116 && 13,824 && 170,832 \\

\end{tabular}
}
\vspace{1mm}
\caption{Details of the number of counterfactual sets after applying different filters in each of the stages.}
\label{table:filtering-stats}
\end{table*}

\subsection{Details of Caption Construction}
\label{app:caption-details}

\phillip{Todo: update this section}
\tile{I updated the list of 156 occupations in Table~\ref{tab:occupations}. Please update the text here accordingly.}

We use a set of 260 occupations in this work, which was collected by combining the occupation lists proposed by \citet{nadeem2020stereoset}, \citet{chuang2023debiasing}, \citet{naik2023social}, and \citet{harrison2023run}.
After applying our three-stage filtering methodology, only 158 occupations remain in our dataset, which are provided in Table~\ref{tab:occupations}.
% For captions which utilized personality traits as the subject, we used the same list of 63 traits as in \citet{naik2023social}, which is provided in Table~\ref{tab:traits}. 
To study bias associated with physical characteristics, we used the following keywords for positive and negative body stereotypes provided in \citet{mei2023bias}: skinny, obese, young, old, tattooed. 

Examples of captions constructed using various prefixes, subjects, and bias attributes are provided in Table~\ref{tab:templateconstruction}. We provide details of the total number of captions and images generated for each subject and attribute pairs in Table~\ref{table:filtering-stats}. The total number of counterfactual sets is determined by the product of the number of prefixes used to construct captions (4), the cardinality of the subject set (i.e., the number of occupations), and 100 (the number of over-generations per set). The number of images per set is determined by the product of the cardinalities of the attribute sets. The total number of generated images is the product of the number of counterfactual sets and the number of images per set.

\section{Additional Discussion of Limitations and Ethical Considerations}
\label{sec:additional-limitations}

\textbf{Limitations} Despite our best efforts, the templates and methodologies we adopt may themselves contain some latent biases which could contribute to the implicit biases exhibited by VLMs. All statements pertaining to gender, race, and occupational attributes or associations should be interpreted only within the context of our experiments. Furthermore, all discussion of social attributes in this work are intended to be interpreted as \textit{perceived}. We are aware that our approach only considers binary classification of genders and does not exhaustively encompass all races, physical characteristics, and occupations, which is due to data limitation rather than value judgements by the authors. The results we present cannot be assumed to generalize to social and demographic terms omitted in our analysis. The labels for the attributes we present in the paper are derived from prior work and were further limited to those which stable diffusion could depict. Our goal is to provide text labels that produce perceived physical differences, but these are not labels we aim to impose on any groups or sub-groups. Similar to \citet{smith2022m}, we recognize there are trade-offs in creating lists of socials attributes. While these lists may not be entirely inclusive, we leverage them for their benefit in identifying and mitigating bias. 
% Any absence of groups or sub-groups should not be considered as disregard or our own bias, but to be treated as a data contraint.
Our study was conducted in English, which limits the generalizability of our findings to other languages. 

\textbf{Ethical Considerations.} With the findings we present in this paper, we aim to increase the understanding of bias in VLMs and probe mitigation strategies. We acknowledge that our work does not encompass all possible social attributes and that our selected categories for gender, race, physical characteristics, and occupations may harbor stereotypes that cannot be assumed to represent their entire groups. Similar to  \citet{hall2023visogender}, we recognize that we may miss intersectional characteristics that constitute a well-accepted image of a person in a specific occupation or belonging to a race. Our aim is that the techniques presented in this work can help reduce various social disparities in VLMs and can be further extended to include more genders, races, occupations and other social characteristics. Continuing these efforts will increase confidence in the ability of VLMs to exhibit fairness with respect to differing social attributes. We understand that the use of a bias reduction strategy without deep understanding of various nuances does not guarantee a foolproof solution in bias elimination, and still may result in VLMs that cause harm and stigmatize certain subsets of individuals. Therefore, debiasing efforts should be further developed prior to wide-spread adoption in sensitive applications.

\begin{figure*}[ht!]
    \centering
    \begin{subfigure}[b]{0.33\textwidth}
    \includegraphics[trim={2mm 2mm 2mm 
    2mm},clip,width=1\columnwidth]{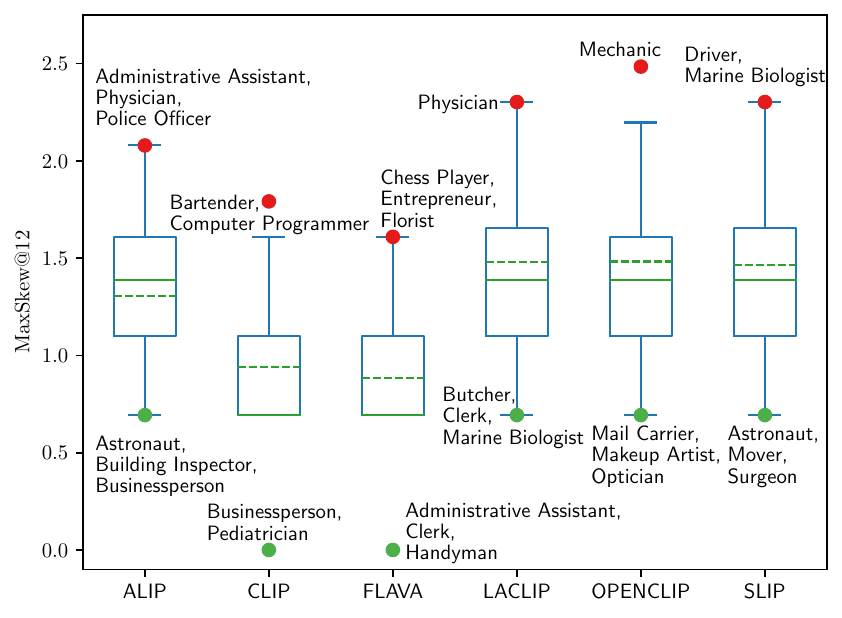}
    \caption{Race-Gender}
    \label{fig:race-gender}
    \end{subfigure}
    \begin{subfigure}[b]{0.33\textwidth}
    \includegraphics[trim={2mm 2mm 2mm 
    2mm},clip,width=1\columnwidth]{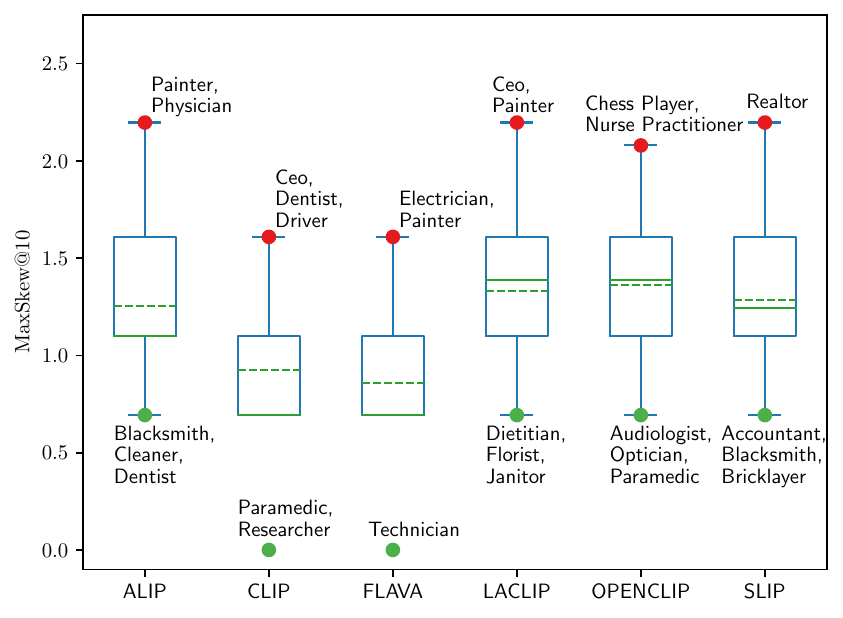}
    \caption{Physical Characteristics-Gender}
    \label{fig:physical-gender}
    \end{subfigure}
    \begin{subfigure}[b]{0.33\textwidth}
    \includegraphics[trim={2mm 2mm 2mm 
    2mm},clip,width=1\columnwidth]{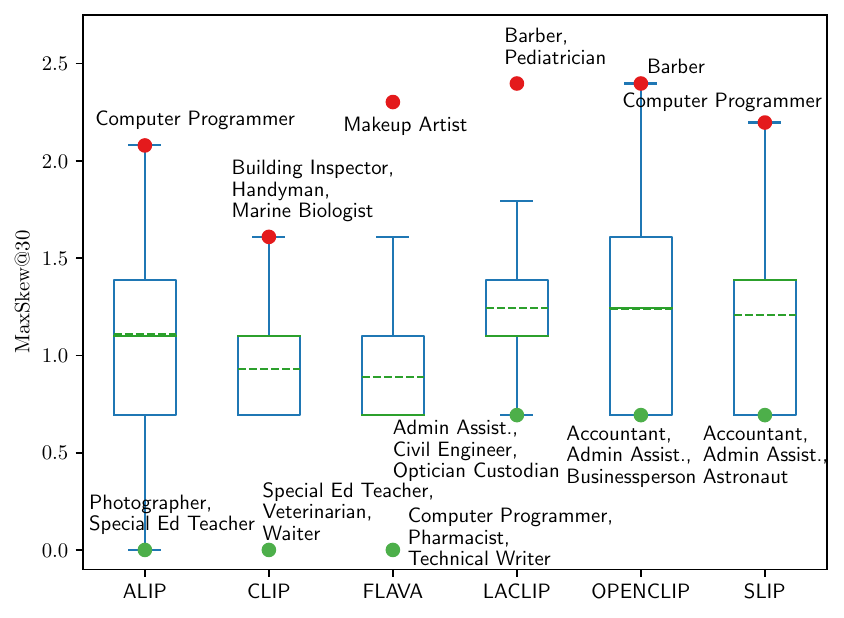}
    \caption{Physical Characteristics-Race}
    \label{fig:physical-race}
    \end{subfigure}
    \caption{
    Distribution of $\textrm{MaxSkew}@K$ measured across occupations for (a) Race-Gender, (b) Physical Characteristics-Gender, and (c) Physical Characteristics-Race intersectional biases after NSFW and Attribute Detectability Filtering. Max (min) values are plotted as red (green) circles with corresponding occupation names
    }
    \label{fig:maxskew-full-filtering}
\end{figure*}

\section{Additional Analysis and Results}

\subsection{NDKL and Bias@K results}
\label{app:ndkl-results}

\begin{table}
\begin{center}
\resizebox{1\columnwidth}{!}
{
\begin{tabular}{lrrrrrrr}
\toprule
   Model &  Mean &  Std &  Min &  25\% &  50\% &  75\% &  Max \\
\midrule
%     ALIP &  0.67 & 0.34 & 0.16 & 0.42 & 0.64 & 0.84 & 1.72 \\
%     CLIP &  0.33 & 0.20 & 0.04 & 0.18 & 0.29 & 0.43 & 1.13 \\
%    FLAVA &  0.31 & 0.19 & 0.05 & 0.15 & 0.28 & 0.44 & 0.92 \\
%   LACLIP &  0.86 & 0.30 & 0.26 & 0.62 & 0.83 & 1.05 & 1.81 \\
% OPENCLIP &  0.89 & 0.36 & 0.20 & 0.68 & 0.85 & 1.08 & 2.39 \\
%     SLIP &  0.85 & 0.36 & 0.13 & 0.55 & 0.85 & 1.09 & 2.00 \\
    ALIP &  0.58 & 0.23 & 0.28 & 0.41 & 0.51 & 0.67 & 1.31 \\
    CLIP &  0.34 & 0.09 & 0.24 & 0.29 & 0.32 & 0.37 & 0.76 \\
   FLAVA &  0.34 & 0.08 & 0.23 & 0.28 & 0.32 & 0.39 & 0.57 \\
  LACLIP &  0.73 & 0.26 & 0.29 & 0.50 & 0.70 & 0.90 & 1.59 \\
OPENCLIP &  0.77 & 0.30 & 0.31 & 0.51 & 0.75 & 0.93 & 2.04 \\
    SLIP &  0.71 & 0.28 & 0.26 & 0.47 & 0.65 & 0.89 & 1.41 \\
\bottomrule
\end{tabular}
}
\end{center}
\caption{Distribution of NDKL scores by model, measured across occupations using the \textbf{race-gender} dataset segment.}
\label{tab:ndkl-race-gender}
\end{table}

\begin{table}
\begin{center}
\resizebox{1\columnwidth}{!}
{
\begin{tabular}{lrrrrrrr}
\toprule
   Model &  Mean &  Std &  Min &  25\% &  50\% &  75\% &  Max \\
\midrule
%     ALIP &  0.37 & 0.27 & 0.04 & 0.15 & 0.33 & 0.53 & 1.18 \\
%     CLIP &  0.22 & 0.14 & 0.04 & 0.11 & 0.20 & 0.30 & 0.60 \\
%    FLAVA &  0.22 & 0.19 & 0.04 & 0.10 & 0.17 & 0.27 & 1.08 \\
%   LACLIP &  0.47 & 0.29 & 0.05 & 0.26 & 0.41 & 0.65 & 1.36 \\
% OPENCLIP &  0.45 & 0.31 & 0.04 & 0.21 & 0.42 & 0.62 & 1.50 \\
%     SLIP &  0.45 & 0.26 & 0.05 & 0.25 & 0.44 & 0.59 & 1.12 \\
    ALIP &  0.52 & 0.25 & 0.21 & 0.33 & 0.43 & 0.73 & 1.05 \\
    CLIP &  0.47 & 0.29 & 0.14 & 0.19 & 0.36 & 0.67 & 1.03 \\
   FLAVA &  0.45 & 0.27 & 0.13 & 0.20 & 0.38 & 0.66 & 0.92 \\
  LACLIP &  0.56 & 0.25 & 0.24 & 0.36 & 0.47 & 0.77 & 1.05 \\
OPENCLIP &  0.54 & 0.25 & 0.21 & 0.32 & 0.46 & 0.71 & 1.05 \\
    SLIP &  0.56 & 0.26 & 0.23 & 0.34 & 0.45 & 0.76 & 1.19 \\
\bottomrule
\end{tabular}
}
\end{center}
\caption{Distribution of NDKL scores by model, measured across occupations using the \textbf{race-physical characteristics} dataset segment.}
\label{tab:ndkl-physical-race}
\end{table}

\begin{table}
\begin{center}
\resizebox{1\columnwidth}{!}
{
\begin{tabular}{lrrrrrrr}
\toprule
   Model &  Mean &  Std &  Min &  25\% &  50\% &  75\% &  Max \\
\midrule
%     ALIP &  0.70 & 0.38 & 0.13 & 0.41 & 0.64 & 0.90 & 2.05 \\
%     CLIP &  0.34 & 0.18 & 0.05 & 0.20 & 0.32 & 0.44 & 0.84 \\
%    FLAVA &  0.27 & 0.15 & 0.04 & 0.16 & 0.25 & 0.35 & 0.71 \\
%   LACLIP &  0.79 & 0.38 & 0.11 & 0.53 & 0.78 & 0.99 & 1.96 \\
% OPENCLIP &  0.79 & 0.32 & 0.28 & 0.53 & 0.78 & 0.96 & 1.76 \\
%     SLIP &  0.73 & 0.32 & 0.12 & 0.51 & 0.70 & 0.93 & 1.66 \\
    ALIP &  0.63 & 0.28 & 0.28 & 0.43 & 0.54 & 0.76 & 1.83 \\
    CLIP &  0.39 & 0.08 & 0.27 & 0.34 & 0.38 & 0.43 & 0.65 \\
   FLAVA &  0.36 & 0.07 & 0.27 & 0.31 & 0.34 & 0.39 & 0.55 \\
  LACLIP &  0.72 & 0.29 & 0.29 & 0.49 & 0.66 & 0.91 & 1.68 \\
OPENCLIP &  0.74 & 0.26 & 0.32 & 0.52 & 0.68 & 0.91 & 1.55 \\
    SLIP &  0.66 & 0.24 & 0.31 & 0.49 & 0.60 & 0.79 & 1.28 \\
\bottomrule
\end{tabular}
}
\end{center}
\caption{Distribution of NDKL scores by model, measured across occupations using the \textbf{gender-physical characteristics} dataset segment.}
\label{tab:ndkl-physical-gender}
\end{table}

\begin{table}
\begin{center}
\resizebox{1\columnwidth}{!}
{
\begin{tabular}{lrrrrrr}
\toprule
Dataset Segment &  ALIP &  CLIP &  FLAVA &  LACLIP &  OPENCLIP &  SLIP \\
\midrule
Physical-Gender &  0.06 & -0.18 &   \textbf{0.02} &    0.51 &      0.50 &  0.40 \\
Race-Gender &  \textbf{0.01} & -0.12 &  -0.02 &    0.47 &      0.50 &  0.41 \\
\bottomrule
\end{tabular}

}
\end{center}
\caption{Bias@K by model, calculated across occupations for dataset segments which include gender attributes.}
\label{tab:bias-at-k}
\end{table}

\begin{table*}
\footnotesize
\begin{center}
% \resizebox{1\columnwidth}{!}
% {
\begin{tabular}{l c c c c c c} %
\toprule
& \multicolumn{2}{c}{\textbf{CLIP}~\cite{radford2021learning}} & \multicolumn{2}{c}{\textbf{ALIP}~\cite{yang2023alip}} & \multicolumn{2}{c}{\textbf{FLAVA} ~\cite{singh2022flava}}  \\
\cmidrule(lr){2-3}
\cmidrule(lr){4-5}
\cmidrule(lr){6-7}
\textbf{Intersectional Bias} & \textbf{Pre-trained} & \textbf{Debiased} & \textbf{Pre-trained} & \textbf{Debiased} & \textbf{Pre-trained} & \textbf{Debiased}\\
\midrule
% $(\textrm{Race}, \textrm{Gender})$ & 0.40 & 0.29 & 0.80 & 0.43 & 0.38 & 0.40\\
% $(\textrm{Physical Char.}, \textrm{Gender})$ & 0.35 & 0.25 & 0.83 & 0.30 & 0.33 & 0.37\\
% $(\textrm{Physical Char.}, \textrm{Race})$ & 0.34 & 0.26 & 0.57 & 0.36 & 0.33 & 0.36\\
$(\textrm{Race}, \textrm{Gender})$ & 0.37 & 0.33 & 0.71 & 0.36 & 0.376 & 0.365 \\
$(\textrm{Physical Char.}, \textrm{Gender})$ & 0.41 & 0.34 & 0.77 & 0.36 & 0.38 & 0.40 \\
$(\textrm{Physical Char.}, \textrm{Race})$ & 0.20 & 0.17 & 0.30 & 0.20 & 0.204 & 0.196 \\
\bottomrule
\end{tabular}
% }
\caption{Mean of NDKL for pre-trained and debiased variants of CLIP, ALIP, and FLAVA, estimated by withholding counterfactual sets for 20\% of the occupations in our dataset.}
\label{tab:intersectional-debias-ndkl}
\end{center}
\end{table*}

\begin{table*}
\footnotesize
\begin{center}
% \resizebox{1\columnwidth}{!}
% {
\begin{tabular}{l c c c c c c} %
\toprule
& \multicolumn{2}{c}{\textbf{CLIP}~\cite{radford2021learning}} & \multicolumn{2}{c}{\textbf{ALIP}~\cite{yang2023alip}} & \multicolumn{2}{c}{\textbf{FLAVA} ~\cite{singh2022flava}}  \\
\cmidrule(lr){2-3}
\cmidrule(lr){4-5}
\cmidrule(lr){6-7}
\textbf{Intersectional Bias} & \textbf{Pre-trained} & \textbf{Debiased} & \textbf{Pre-trained} & \textbf{Debiased} & \textbf{Pre-trained} & \textbf{Debiased}\\
\midrule
$(\textrm{Race}, \textrm{Gender})$ & -0.19 & 0.13 & 0.15 & 0.20 & -0.05 & 0.10 \\
$(\textrm{Physical Char.}, \textrm{Gender})$ & -0.22 & 0.06 & 0.09 & 0.03 & -0.09 & 0.02 \\
\bottomrule
\end{tabular}
% }
\caption{Bias@K for pre-trained and debiased variants of CLIP, ALIP, and FLAVA, estimated by withholding counterfactual sets for 20\% of the occupations in our dataset.}
\label{tab:intersectional-debias-bias-at-k}
\end{center}
\end{table*}

In addition to MaxSkew, we also calculated the Bias@K and Normalized Discounted Kullback-Leibler Divergence (NDKL) metrics for our bias probing \& mitigation experiments. See \citet{geyik2019fairness} for a detailed description of NDKL and \citet{wang2021gender} for details of Bias@K.

We estimate NDKL by summing over ranked lists of size $\{1,2,..,K^{2}\}$, where $K=|A_{1}| \times |A_{2}|$. Tables~\ref{tab:ndkl-race-gender},~\ref{tab:ndkl-physical-race}, and~\ref{tab:ndkl-physical-gender} provide the distribution of NDKL scores for each model, measured across occupations for the three segments of SocialCounterfactuals. Consistent with our analysis of MaxSkew@K (Section~\ref{sec:results}), CLIP and FLAVA generally have the lowest NDKL scores. Notably, all models exhibit values well above the ideal case of 0 for this metric. 

Table~\ref{tab:bias-at-k} provides Bias@K for our two dataset segments which include gender attributes, where we set $K=|A_{1}| \times |A_{2}|$ as in our MaxSkew@K and NDKL evaluations. Bias@K is a measure of marginal gender bias; a value of 0 indicates that both genders are represented equally in retrieval results. Positive values indicate that males are represented more than females, while negative values indicate that females are represented more than males. ALIP and FLAVA exhibit the least amount of gender bias overall. CLIP demonstrates some bias towards female images, while LACLIP, OpenCLIP, and SLIP all exhibit a strong bias for male images.

% On model debiasing (Table~\ref{tab:intersectional-debias}), we observe an average relative improvement of 5\% with CLIP, ALIP, and FLAVA for both metrics.
Tables~\ref{tab:intersectional-debias-ndkl} and~\ref{tab:intersectional-debias-bias-at-k} provide NDKL and Bias@K results (respectively) for models which were evaluated in our debiasing experiments (Section~\ref{sec:mitigating}). Because Bias@K only measures gender bias, we only calculate it for dataset segments which include gender attributes. 

\subsection{Preliminary evaluations using counterfactuals with three intersectional attributes}
\label{app:three-attributes}

Our dataset generation approach can incorporate greater combinations of attributes by introducing additional attribute sets to our caption templates (Section~\ref{sec:captions}). To investigate this, we produced 16k images spanning 266 counterfactual image sets with three intersectional attributes in each caption (physical characteristics, gender, and race). The CLIP MaxSkew@60 score for three-attribute intersectional bias on this set of images is 0.693, which reveals even greater bias than when measuring various pairs of two-attribute intersectional bias (which have a MaxSkew@60 of 0.567-0.59 on the same images). 

We believe this preliminary result points to a promising direction for future studies, which could apply our approach to investigate bias across a greater number of intersectional attributes.
To the best of our knowledge, our work is the first to address the problems of probing and mitigating intersectional bias in vision-language models. While we focused on pairs of attributes in our paper, we believe this nonetheless represents a significant contribution considering the lack of prior work addressing intersectional social biases. 

\subsection{FID and IS evaluation of SocialCounterfactuals and other datasets}
\label{app:fid-and-is}

We computed Inception Score (IS) and FID relative to ImageNet and LAION-2B for our dataset, VisoGender, and PATA. As shown in the table below, our dataset outperforms both VisoGender and PATA (which consist of real-world images for bias evaluation) across all metrics.
\begin{table}[h!]
% \footnotesize
\begin{center}
\resizebox{1\columnwidth}{!}
{
\begin{tabular}{l c c c} %
\toprule
& \textbf{FID (ImageNet) $\downarrow$} & \textbf{FID (LAION-2B) $\downarrow$} & \textbf{IS $\uparrow$} \\
\midrule
VisoGender & 130.44 & 113.45 & 6.2\\
PATA & 116.29 & 98.34 & 11.3\\
SocialCounterfactuals & \textbf{106.83} & \textbf{89.91} & \textbf{12.07}\\
\bottomrule
\end{tabular}
}
% \caption{.}
\label{tab:fid_is}
\end{center}
\end{table}

\subsection{Additional details and results for bias probing experiments}
\label{app:probing-additional-reults}

Our bias probing analysis presented in Section~\ref{sec:results} utilized our SocialCounterfactuals dataset before NSFW and Attribute Detectability Filtering was applied. For completeness, in this section we provide the same analysis on the SocialCounterfactuals dataset after NSFW and Attribute Detectability Filtering. 

\begin{figure}[t!]
    \centering
    \includegraphics[trim={2mm 2mm 2mm 
    2mm},clip,width=1\columnwidth]{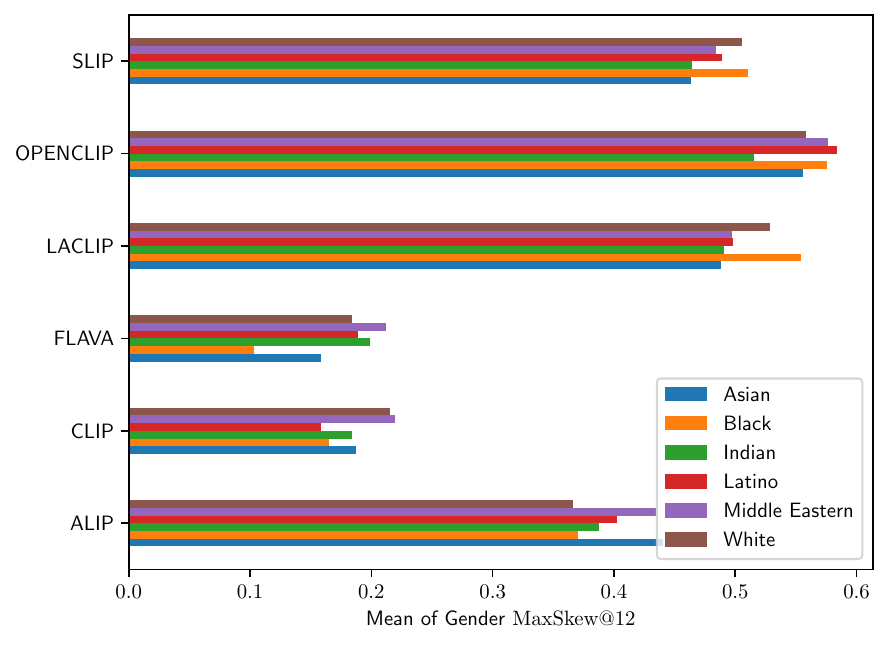}
    \caption{
    Mean of (marginal) gender $\textrm{MaxSkew}@K$ measured across occupations for different races after NSFW and Attribute Detectability Filtering.
    }
    \label{fig:maxskew-by-race-full-filtering}
\end{figure}

\begin{figure*}[t!]
    \centering
    \includegraphics[trim={2mm 2mm 2mm 
    2mm},clip,width=1\textwidth]{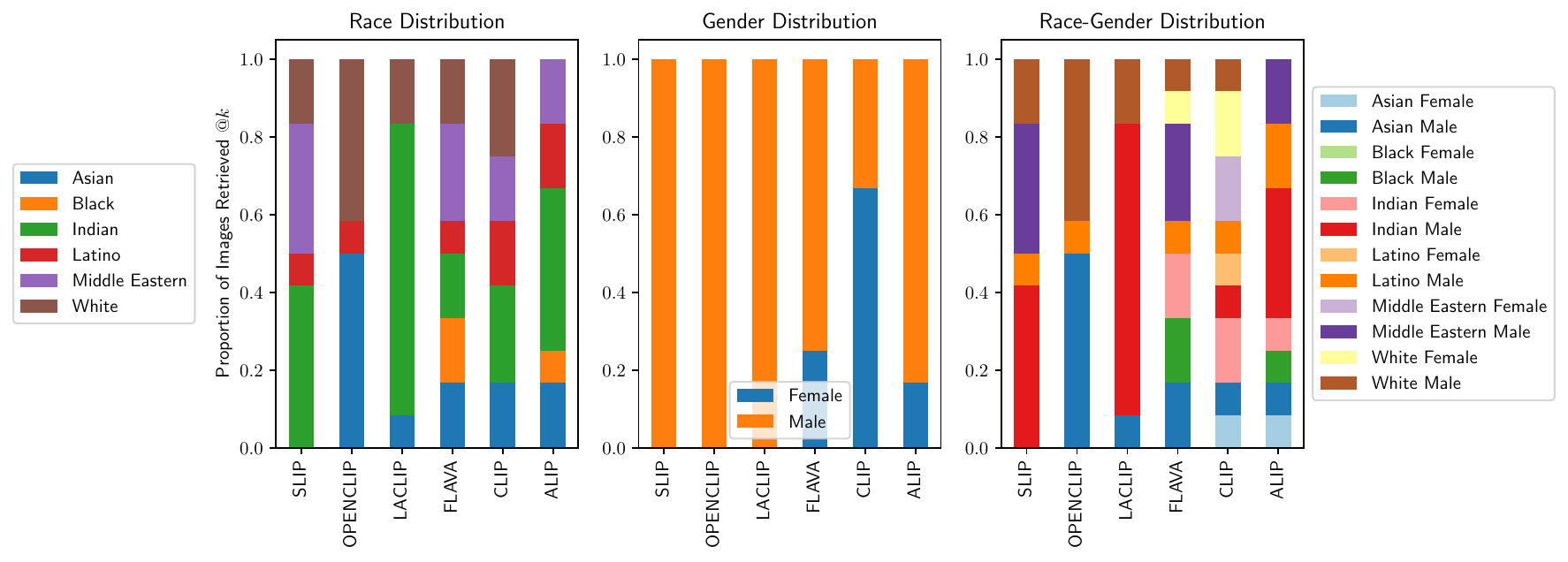}
    \caption{
    Proportion of images retrieved $@k=12$ after NSFW and Attribute Detectability Filtering using neutral prompts for the \textbf{Doctor} occupation, broken down by race \& gender attributes.
    }
    \label{fig:doctor-full-filtering}
\end{figure*}

\begin{figure*}[t!]
    \centering
    \includegraphics[trim={2mm 2mm 2mm 
    2mm},clip,width=1\textwidth]{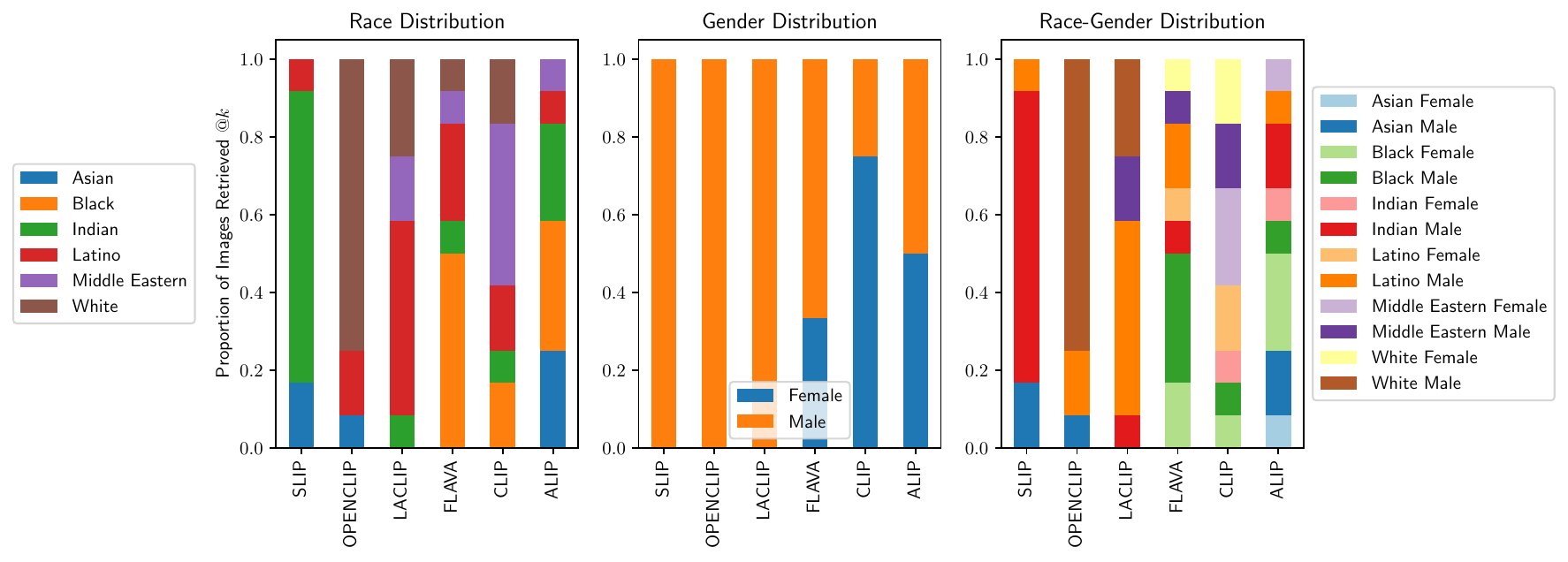}
    \caption{
    Proportion of images retrieved $@k=12$ after NSFW and Attribute Detectability Filtering using neutral prompts for the \textbf{Software Developer} occupation, broken down by race \& gender attributes.
    }
    \label{fig:race-gender-software-developer}
\end{figure*}

\begin{figure*}[t!]
    \centering
    \includegraphics[trim={2mm 2mm 2mm 
    2mm},clip,width=1\textwidth]{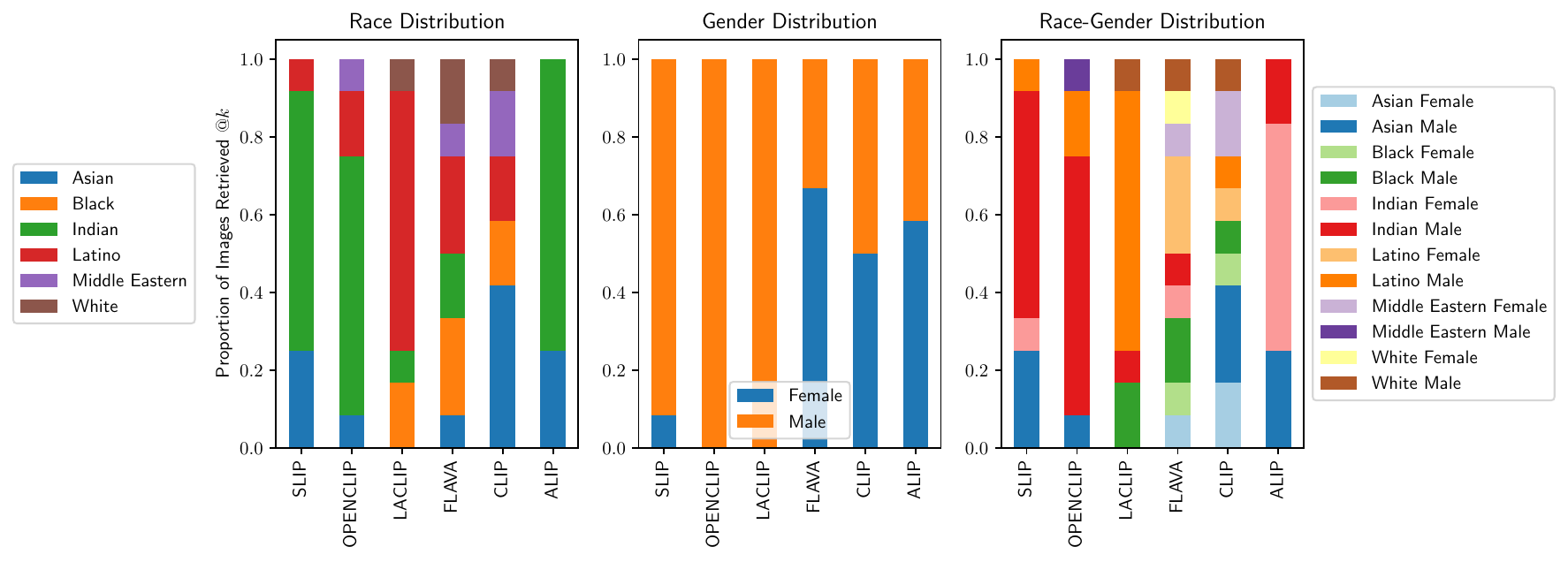}
    \caption{
    Proportion of images retrieved $@k=12$ after NSFW and Attribute Detectability Filtering using neutral prompts for the \textbf{Construction Worker} occupation, broken down by race \& gender attributes.
    }
    \label{fig:race-gender-construction-worker}
\end{figure*}

\begin{figure*}[t!]
    \centering
    \includegraphics[trim={2mm 2mm 2mm 
    6mm},clip,width=1\textwidth]{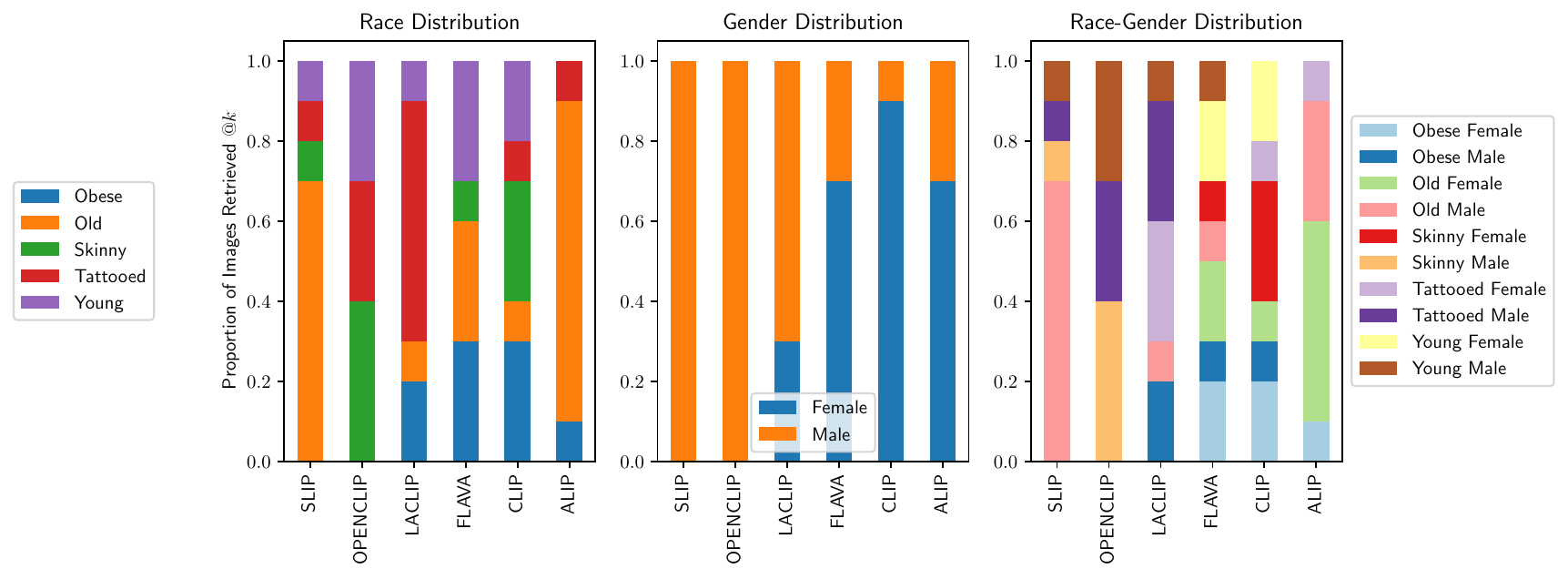}
    \caption{
    Proportion of images retrieved $@k=10$ after NSFW and Attribute Detectability Filtering using neutral prompts for the \textbf{Entrepreneur} occupation, broken down by gender \& physical characteristics.
    }
    \label{fig:physical-gender-entrepreneur}
\end{figure*}

\begin{figure*}[t!]
    \centering
    \includegraphics[trim={2mm 2mm 2mm 
    6mm},clip,width=1\textwidth]{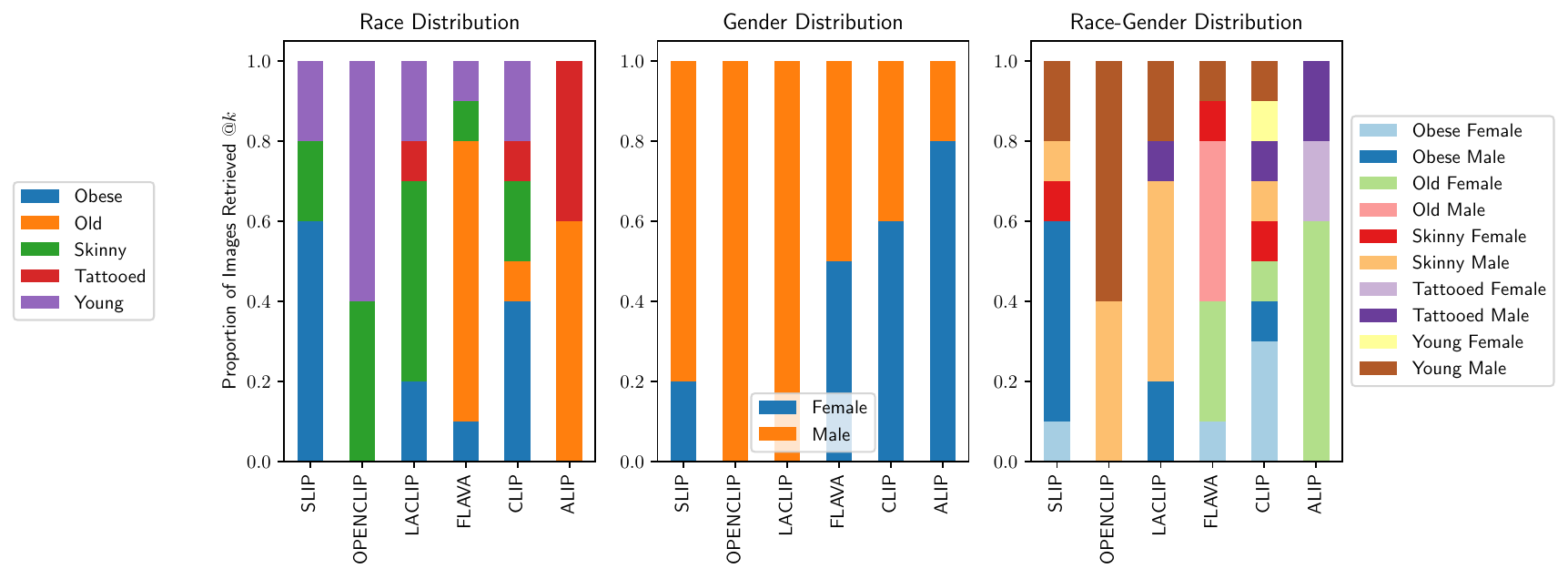}
    \caption{
    Proportion of images retrieved $@k=10$ after NSFW and Attribute Detectability Filtering using neutral prompts for the \textbf{Technical Writer} occupation, broken down by gender \& physical characteristics.
    }
    \label{fig:physical-gender-technical-writer}
\end{figure*}

\begin{figure*}[t!]
    \centering
    \includegraphics[trim={2mm 2mm 2mm 
    6mm},clip,width=1\textwidth]{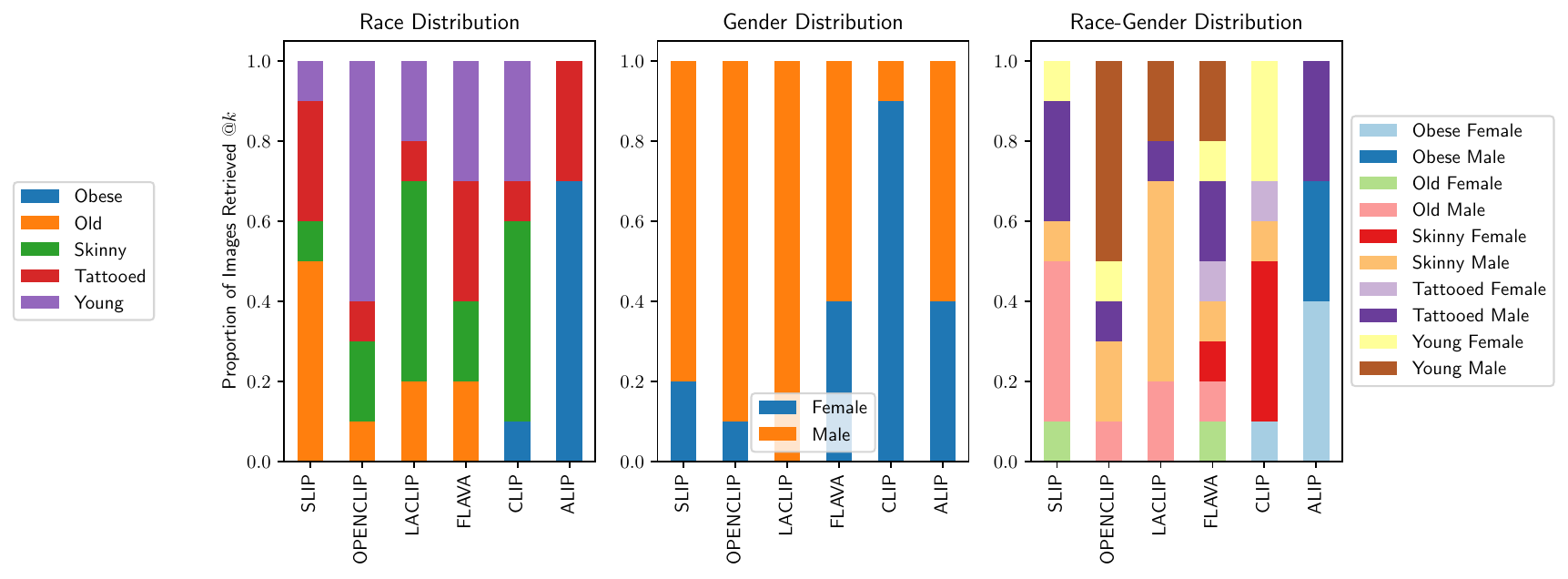}
    \caption{
    Proportion of images retrieved $@k=10$ after NSFW and Attribute Detectability Filtering using neutral prompts for the \textbf{Salesperson} occupation, broken down by gender \& physical characteristics.
    }
    \label{fig:physical-gender-salesperson}
\end{figure*}

Figure~\ref{fig:maxskew-full-filtering}  provides boxplots of the intersectional bias $\textrm{MaxSkew}@K$ distribution for each VLM, measured across occupations separately using the three segments of our dataset after NSFW and Attribute Detectability Filtering. Overall we find that these distributions are largely similar to those described previously in Figure~\ref{fig:maxskew}. Similarly, the marginal gender skewness depicted in Figure~\ref{fig:maxskew-by-race-full-filtering} and retrieval proportions for the `Doctor' profession illustrated in Figure~\ref{fig:doctor-full-filtering} (post NSFW and Attribute Detectability Filtering) reflect largely the same trends as those discussed previously in Section~\ref{sec:results} for Figures~\ref{fig:maxskew-by-race} and~\ref{fig:doctor}. We also provide retrieval proportions for other occupations in Figures~\ref{fig:race-gender-software-developer} to~\ref{fig:physical-gender-salesperson}.

\begin{table*}
\centering
\small
% \resizebox{\textwidth}{!}{%
\begin{tabular}{p{11.25cm}}
\toprule
\textbf{Occupations} \\
\midrule
academic, accountant, administrative assistant, analyst, architect, army, artist, assistant, astronaut, athlete, attorney, audiologist, auditor, author, baker, banker, barber, bartender, biologist, blacksmith, boxer, bricklayer, broker, building inspector, bus driver, businessperson, butcher, carpenter, cashier, ceo, chef, chemist, chess player, chief, chief executive officer, childcare worker, civil engineer, civil servant, cleaner, clerk, coach, comedian, commander, composer, computer programmer, construction worker, consultant, cook, crane operator, customer service representative, dancer, delivery man, dentist, designer, detective, dietitian, dj, doctor, driver, economist, editor, electrician, engineer, entrepreneur, farmer, firefighter, florist, football player, graphic designer, guard, guitarist, hairdresser, handball player, handyman, housekeeper, janitor, judge, lab tech, laborer, lawyer, librarian, magician, mail carrier, makeup artist, manager, marine biologist, mathematician, mechanic, model, mover, musician, nurse, nurse practitioner, nutritionist, opera singer, optician, optician custodian, painter, paramedic, pastry chef, pediatrician, pensioner, pharmacist, photographer, physician, physicist, pianist, pilot, plumber, poet, police officer, policeman, pr person, priest, primary school teacher, prisoner, producer, professor, psychologist, real estate developer, real estate agent, realtor, receptionist, recruiter, reporter, researcher, roofer, sailor, salesperson, scientist, secretary, security guard, sheriff, software developer, soldier, special ed teacher, statistician, supervisor, surgeon, surveyor, swimmer, tailor, teacher, technical writer, technician, telemarketer, tennis player, therapist, tour guide, umpire, veterinarian, videographer, waiter, web developer, writer, zoologist \\
\bottomrule
\end{tabular}
% }
\caption{Full list of occupations which remain in our dataset after filtering.}
\label{tab:occupations}
\end{table*}

% \begin{table*}[ht]
% \centering
% \small
% % \resizebox{\textwidth}{!}{%
% \begin{tabular}{p{11.25cm}}
% \toprule
% \textbf{Physical characteristics} \\
% \midrule
% skinny, obese, young, old, tattooed \\
% \bottomrule
% \end{tabular}
% % }
% \caption{Full list of physical characteristics.}
% \label{tab:physicaltraits}
% \end{table*}

\begin{table*}
\footnotesize
	\begin{center}
		% \resizebox{1\textwidth}{!}
		{
		\begin{tabular}{l l l l l} %
	    \hline
         \textbf{Template\Tstrut\Bstrut} & \textbf{Attribute-1\Tstrut\Bstrut} & \textbf{Attribute-2\Tstrut\Bstrut} & \textbf{Subject\Tstrut\Bstrut} & \textbf{Caption\Tstrut\Bstrut}\\
        \hline
A/An <race> <gender> <occupation> \Tstrut\Bstrut & Indian \Tstrut\Bstrut & female \Tstrut\Bstrut & nurse \Tstrut\Bstrut & An Indian female nurse \Tstrut\Bstrut \\
A photo of <race> <gender> <occupation>  & White & male  & electrician & A photo of a White male electrician \\
A picture of <physical char.> <race> <occupation> & young & Latino & web developer & A picture of a young Latino web developer \\
An image of <physical char.> <gender> <occupation> \Bstrut & tattooed \Bstrut & male \Bstrut & barber \Bstrut & An image of a tattooed male barber \Bstrut \\
 \hline
		\end{tabular}}
		\caption{Examples of captions constructed from templates}
        \label{tab:templateconstruction}
	\end{center}
\end{table*}

\end{document}